\pdfoutput=1
\documentclass[11pt]{article}

\usepackage[]{acl}

\usepackage{times}
\usepackage{latexsym}
\usepackage{wrapfig}
\usepackage{xcolor,pifont}
\usepackage{graphicx}
\usepackage{quoting}
\usepackage{floatrow}
\newcommand*\colourcheck[1]{%
  \expandafter\newcommand\csname #1check\endcsname{\textcolor{#1}{\ding{52}}}%
}
\colourcheck{blue}
\colourcheck{green} 
\colourcheck{red}

\usepackage{soul}

\usepackage{amssymb}% http://ctan.org/pkg/amssymb
\usepackage{pifont}% http://ctan.org/pkg/pifont
\newcommand{\xmark}{\textcolor{purple}{\ding{55}}}
\newcommand{\emark}{\textcolor{blue}{\ding{41}}}

\usepackage{graphicx}
\usepackage{tikz}
\usepackage{forest}
\usetikzlibrary{trees,positioning,shapes,shadows,arrows.meta}

\usepackage[T1]{fontenc}
\usepackage[utf8]{inputenc}

\usepackage{microtype}
\usepackage{amsmath}

\title{Distractor Generation in Multiple-Choice Tasks: A Survey of Methods, Datasets, and Evaluation}

\author{Elaf Alhazmi$^{1,}$$^3$, Quan Z. Sheng$^1$, Wei Emma Zhang$^2$,  Munazza Zaib$^1$, Ahoud Alhazmi$^3$ \\[2mm]
      $^1$School of Computing, Macquarie University, Australia\\
      $^2$School of Computer and Mathematical Sciences, The University of Adelaide, Australia\\
      $^3$College of Engineering and Computing in Al-Lith, Umm Al-Qura University, Saudi Arabia\\
       \texttt{elaf.alhazmi@hdr.mq.edu.au}, \texttt{\{eafhazmi, aafhazmi\}@uqu.edu.sa}, \\
       \texttt{\{michael.sheng, munazza-zaib\}@mq.edu.au}, \\ \texttt{wei.e.zhang@adelaide.edu.au} 
 }

\begin{document}

\maketitle

\begin{abstract}
\frenchspacing
The distractor generation task focuses on generating incorrect but plausible options for objective questions such as fill-in-the-blank and multiple-choice questions. 
This task is widely utilized in educational settings across various domains and subjects. The effectiveness of these questions in assessments relies on the quality of the distractors, as they challenge examinees to select the correct answer from a set of misleading options. The evolution of artificial intelligence (AI) has transitioned the task from traditional methods to the use of neural networks and pre-trained language models. This shift has established new benchmarks and expanded the use of advanced deep learning methods in generating distractors. This survey explores distractor generation tasks, datasets, methods, and current evaluation metrics for English objective questions, covering both text-based and multi-modal domains. 
It also evaluates existing AI models and benchmarks and discusses potential future research directions\footnote{Resources are available at \url{https://github.com/Distractor-Generation/DG_Survey}.}.
\end{abstract}

\section{Introduction}
\frenchspacing
Objective questions \cite{das2021automatic} such as fill-in-the-blank and multiple-choice questions require an examinee to select one valid answer from a set of invalid options \cite{kurdi2020systematic}. 
These types of questions contribute to fair assessment across various domains (e.g., Science \cite{liang-etal-2018-distractor}, English \cite{panda-etal-2022-automatic}, Math \cite{mcnichols2023exploring}, and Medicine \cite{ha-yaneva-2018-automatic}).
They are also beneficial for educators in assessing large capacity of students with unbiased results \cite{ch2018automatic}.
However, creating objective questions manually is a laborious task, as it requires selecting plausible false options, known as {\em distractors}, that can effectively confuse the examinee.

Distractor Generation (DG) \cite{chen2022survey} is the process of generating an erroneous plausible option in objective questions.
In automatic generation, various approaches are utilized, including retrieving-based methods \cite{ren2021knowledge}, learning-based approach \cite{liang-etal-2018-distractor} that ranks options according to a set of features, deep neural networks \cite{maurya2020learning}, and pre-trained language models \cite{chiang-etal-2022-cdgp}. 
These methods are applied to distractors in fill-in-the-blank \cite{wang-etal-2023-distractor} and multiple-choice questions, including question answering \cite{bitew2023distractor}, reading comprehension \cite{gao2019generating} and multi-modal \cite{Lu_2022_CVPR} domains. 

Despite the emerging interest in the DG research, there is no literature review in this field, to the best of our knowledge.  Existing relevant surveys focus on generating multiple-choice questions \cite{ch2018automatic,kurdi2020systematic,das2021automatic,zhang2021review} without discussing DG tasks. 
A recent work \cite{chen2022survey} discussed DG as a subtask of natural language generation (NLG) in the text abbreviation tasks, rather than a subtask in objective questions. 
We aim to fill the gap and conduct the first survey for DG in objective type of questions.
To this end, we collected over 100 high-quality papers from top conferences such as ACL, AAAI, IJCAI, ICLR, EMNLP, NAACL, COLING, and AIED
and journals such as ACM Computing Surveys, ACM Transactions on Information System, IEEE Transactions on Learning Technologies and IEEE/ACM Transactions on Audio, Speech, and Language Processing.

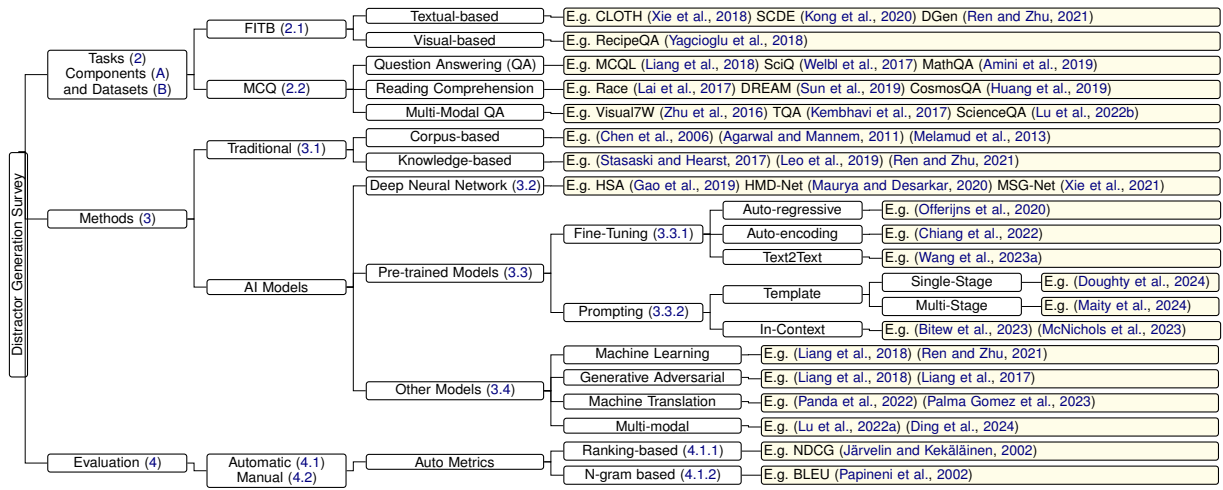
\begin{figure*}
    \centering
    \tikzset{
        basic/.style  = {draw, ultra thick, text width=7 cm, align=center, font=\sffamily\huge, rounded corners},
        root/.style   = {basic, ultra thick=2pt, align=center, text width=30em},
        onode/.style  = {basic, ultra thick=2pt, align=center, text width=9cm},
        tnode/.style  = {basic, ultra thick, align=left, fill=yellow!10, text width=34cm},
        mnode/.style  = {basic, ultra thick, align=left, fill=yellow!10, text width=23.6cm},
        snode/.style  = {basic, ultra thick, align=left, fill=yellow!10, text width=9cm},
        lnode/.style  = {basic, ultra thick, align=left, fill=yellow!10, text width=25cm},
        xnode/.style  = {basic, ultra thick, align=left, fill=yellow!10, text width=17.4cm},  
        wnode/.style  = {basic, ultra thick, align=left, fill=pink!10!blue!80!red!10, text width=8em},
        edge from parent/.style={draw=black, ultra thick, edge from parent fork right}
    }
    \resizebox{\textwidth}{!}{
    \begin{forest} for tree={
        grow=east,
        growth parent anchor=west,
        parent anchor=east,
        child anchor=west,
        % l sep=15mm,
        % s sep = 5mm,
        edge path={\noexpand\path[\forestoption{edge},-, >={latex}] 
             (!u.parent anchor) -- +(10pt,0pt) |-  (.child anchor) 
             \forestoption{edge label};}
    }
    [Distractor Generation Survey, root, rotate=90, l sep=10mm,
        [Evaluation (\ref{sec:evaluation}), basic, l sep=10mm 
            [Automatic (\ref{sec:automatic_E}) Manual (\ref{sec:human_E}), basic , l sep=10mm
                [Auto Metrics, onode, l sep=10mm
                    [N-gram based (\ref{sec:NLG}), onode, l sep=10mm
                        [E.g. BLEU \cite{papineni-etal-2002-bleu}, mnode]
                    ]
                    [Ranking-based (\ref{sec:ranking}), onode , l sep=10mm
                        [E.g. NDCG \cite{jarvelin2002cumulated}, mnode]
                    ]
                ]     
            ]
        ] 
        [Methods (\ref{sec: Methods}), basic, l sep=10mm
            [AI Models, basic , l sep=10mm
                [Other Models (\ref{sec:Others}), onode , l sep=10mm
                    [Multi-modal, onode, l sep=10mm
                        [E.g. \cite{Lu_2022_CVPR} \cite{ding-etal-2024-learn-question}, mnode]
                    ]
                    [Machine Translation, onode, l sep=10mm
                        [E.g. \cite{panda-etal-2022-automatic} \cite{palma-gomez-etal-2023-using}, mnode]
                    ]
                    [Generative Adversarial, onode, l sep=10mm
                        [E.g. \cite{liang-etal-2018-distractor} \cite{liang2017distractor}, mnode]
                    ]
                    [Machine Learning, onode, l sep=10mm
                        [E.g. \cite{liang-etal-2018-distractor} \cite{ren2021knowledge}, mnode]   
                    ]
                ]
                [Pre-trained Models (\ref{sec:PLMs}), onode, l sep=10mm
                    [Prompting (\ref{sec:Prompting}), basic, l sep=10mm
                        [In-Context, basic, l sep=10mm
                            [E.g. \cite{bitew2023distractor} \cite{mcnichols2023exploring} , xnode]
                        ]
                        [Template, basic , l sep=10mm
                            [Multi-Stage, basic , l sep=10mm
                                [E.g. \cite{maity2024novel}, snode]
                            ]
                            [Single-Stage, basic , l sep=10mm
                                [E.g. \cite{doughty2024comparative}, snode]
                            ]  
                        ]
                    ] 
                    [Fine-Tuning (\ref{sec:Fine-Tuning}), basic, l sep=10mm
                         [Text2Text, basic, l sep=10mm
                            [E.g. \cite{wang-etal-2023-distractor} , xnode]
                        ]
                        [Auto-encoding, basic, l sep=10mm
                            [E.g. \cite{chiang-etal-2022-cdgp} , xnode]
                        ]
                        [Auto-regressive, basic, l sep=10mm
                            [E.g. \cite{offerijns2020better} , xnode]
                        ]
                        % [E.g. BDG \cite{chung-etal-2020-bert} CDGP \cite{chiang-etal-2022-cdgp}, tnode]
                    ]
                ]
                [Deep Neural Network (\ref{sec:Seq2Seq}), onode, l sep=10mm
                    [E.g. HSA \cite{gao2019generating} HMD-Net \cite{maurya2020learning} MSG-Net \cite{xie2021diverse}, tnode]
                ] 
            ]
            [Traditional (\ref{sec:traditional}), basic , l sep=10mm
                [Knowledge-based, onode , l sep=10mm
                    [E.g. \cite{stasaski-hearst-2017-multiple} \cite{leo2019ontology} \cite{ren2021knowledge}, tnode]
                ]
                [Corpus-based, onode, l sep=10mm
                    [E.g. \cite{chen-etal-2006-fast} \cite{agarwal-mannem-2011-automatic} \cite{melamud-etal-2013-two} , tnode]
                ]
            ]
        ] 
        [Tasks (\ref{sec:distractor_generation_tasks}) \\ Components (\ref{sec:dataset_analysis}) and Datasets (\ref{sec:analysis}), basic, l sep=10mm
            % [Tasks, basic
                [MCQ (\ref{sec:MC-DG}), basic , l sep=10mm
                    [Multi-Modal QA, onode , l sep=10mm
                        [E.g. Visual7W \cite{zhu2016visual7w} TQA \cite{kembhavi2017you} ScienceQA \cite{lu2022learn}, tnode]
                    ]
                    [Reading Comprehension, onode, l sep=10mm
                        [E.g. Race \cite{lai-etal-2017-race} DREAM \cite{sun-etal-2019-dream} CosmosQA \cite{huang-etal-2019-cosmos}, tnode]
                    ]
                    [Question Answering (QA), onode, l sep=10mm
                        [E.g. MCQL \cite{liang-etal-2018-distractor} SciQ \cite{welbl-etal-2017-crowdsourcing} MathQA \cite{amini-etal-2019-mathqa}, tnode]
                    ]  
                ]
                [FITB (\ref{sec:FITB}), basic, l sep=10mm
                    [Visual-based, onode,  l sep=10mm
                        [E.g. RecipeQA \cite{yagcioglu-etal-2018-recipeqa}, tnode]
                    ]
                    [Textual-based, onode, l sep=10mm
                        [E.g. CLOTH \cite{xie-etal-2018-large} SCDE \cite{kong-etal-2020-scde} DGen \cite{ren2021knowledge}, tnode]
                    ]
                ] 
            % ]
        ]
    ]
    \end{forest}
    }
    \caption{The Survey Tree for DG. The tasks are fill-in-the-blank (FITB) and multiple-choice question (MCQ).}
    \label{fig:lit_surv}
    \vspace{-2mm}
\end{figure*}

This paper explores English DG and provides a comprehensive understanding of this research area. Figure \ref{fig:lit_surv} illustrates the DG survey tree. 
Our main contributions include: 
%(1)
conducting a detailed review of the DG tasks (Sec.~\ref{sec:distractor_generation_tasks}), related datasets, and methods (Sec.~\ref{sec: Methods});
%(2)
summarizing the evaluation metrics (Sec.~\ref{sec:evaluation}); 
%(3)
discussing the main findings, including the analysis of AI models
and benchmarks (Sec. \ref{sec:Findings});  
%(4)
discussing future research directions (Sec.~\ref{sec:Future_work}); 
%(5)
and providing concluding remarks (Sec.~\ref{sec:conclusion}).

\section{Tasks - Distractor Generation} \label{sec:distractor_generation_tasks}
\frenchspacing
The tasks are categorized into (i) {\em fill-in-the-blank} and (ii) {\em multiple-choice questions}. 
Table \ref{tab:Multiple Choice Datasets} summarizes the available datasets\footnote{We count sub-datasets (CLOTH, RACE, ARC, MCTest).} and categorizes each dataset based on DG tasks. 
A discussion and analysis of the components and datasets are outlined in  Appendix~\ref{sec:dataset_analysis} and Appendix~\ref{sec:analysis}, respectively.

\begin{table*}[tb!]
\centering
% \footnotesize
\caption{Multiple-Choice Datasets. \textbf{K} : thousand, \textbf{M} : million,  \greencheck : public available, \emark : available upon request.}
\label{tab:Multiple Choice Datasets}
\resizebox{\textwidth}{!}{
\begin{tabular}{lccccccc}
\hline\hline
Dataset     & Task     & Domain & Source & Creation & Corpus (C) & C.Unit  & Availability \\\hline
CLOTH \cite{xie-etal-2018-large} 
& FITB      & English exam & Educational  & Expert & 7,131 & Passage & \greencheck \\
CLOTH-M \cite{xie-etal-2018-large} 
& FITB      & English exam & Educational  & Expert & 3,031 & Passage  & \greencheck \\
CLOTH-H \cite{xie-etal-2018-large}  
& FITB      & English exam & Educational  & Expert & 4,100 & Passage  & \greencheck \\
SCDE \cite{kong-etal-2020-scde}     
& FITB & English exam & Educational  & Expert & 5,959 & Passage  & \emark \\
DGen \cite{ren2021knowledge} 
& FITB     & Multi-domain & Multi & Auto & 2,880 & Sentence  & \greencheck \\
CELA \cite{zhang-etal-2023-cloze}   
& FITB    & English exam & Multi & Auto & 150 & Passage  & \greencheck \\
\hline
SciQ \cite{welbl-etal-2017-crowdsourcing} 
& MC-QA     & Science exam & Educational & Crowd & 28& Book  & \greencheck \\
AQUA-RAT \cite{ling-etal-2017-program}
& MC-QA    & Math problem  & Web  & Crowd  &  97,975 & Problem & \greencheck\\
OpenBookQA \cite{mihaylov-etal-2018-suit}
& MC-QA      & Science exam  & Educational \& WorldTree & Crowd &1,326 & WorldTree fact  & \greencheck\\
ARC \cite{clark2018think} 
&MC-QA      & Science exam & Educational \& Web  & Expert  & 14M & Sentence  & \greencheck \\
ARC-Challange \cite{clark2018think} 
& MC-QA      & Science exam  & Educational \& Web & Expert & 14M & Sentence  & \greencheck \\
ARC-Easy \cite{clark2018think} 
& MC-QA      & Science exam  &  Educational \& Web & Expert  & 14M & Sentence   & \greencheck\\
MCQL \cite{liang-etal-2018-distractor} 
&MC-QA     & Science exam  & Educational  \& Web & Crawl  & 7,116 & Query  & \greencheck \\
CommonSenseQA \cite{talmor-etal-2019-commonsenseqa} 
& MC-QA     & Narrative & ConceptNet & Crowd  & 236,208 & ConceptNet Triplets  & \greencheck \\
MathQA \cite{amini-etal-2019-mathqa} 
& MC-QA     & Math problem   & Web  & Crowd  &  37,297   & Problem   & \greencheck \\
QASC \cite{khot2020qasc}
& MC-QA     & Science exam  & Educational \& WorldTree & Crowd & 17M & Sentence  & \greencheck \\
MedMCQA\cite{pal2022medmcqa} 
& MC-QA     &  Medicine exam & Educational &  Expert & 2.4K  & Topics  & \greencheck \\
Televic \cite{bitew2022learning} 
& MC-QA     & Multi-domain & Educational & Expert &  62,858 & Query & \greencheck \\
EduQG \cite{hadifar2023eduqg}
& MC-QA    & Education & Educational & Expert & 13/283 & Book/Chapter & \greencheck \\
\hline
ChildrenBookTest \cite{hill2016goldilocks} 
& MC-RC    & Story & Project Gutenberg  & Auto & 108 & Book & \greencheck\\
Who Did What \cite{onishi-etal-2016-large} 
& MC-RC    & News & Gigaword  & Auto & 10,507 & Book  & \emark\\
MCTest-160 \cite{richardson-etal-2013-mctest}      
& MC-RC & Children story & Fiction   & Crowd & 160 & Story  & \greencheck \\
MCTest-500 \cite{richardson-etal-2013-mctest}      
& MC-RC & Children story & Fiction    & Crowd & 500 & Story  & \greencheck\\
RACE \cite{lai-etal-2017-race}    
& MC-RC &  English exam    & Educational     & Expert &27,933 & Passage  & \greencheck \\
RACE-M \cite{lai-etal-2017-race}
& MC-RC & English exam  & Educational     & Expert & 7,139 & Passage & \greencheck \\
RACE-H \cite{lai-etal-2017-race} 
& MC-RC & English exam & Educational        & Expert &20,784 & Passage  & \greencheck \\
RACE-C \cite{liang2019new} 
& MC-RC & English exam  & Educational   & Expert &4,275& Passage  & \greencheck \\
DREAM \cite{sun-etal-2019-dream} 
& MC-RC & English exam    & Educational    & Expert  & 6,444& Dialogue  & \greencheck\\
CosmosQA \cite{huang-etal-2019-cosmos} 
& MC-RC & Narratives  & Blog    & Crowd & 21,866& Narrative  & \greencheck\\
ReClor \cite{yu2019reclor} 
& MC-RC & Standard exam & Educational  & Expert & 6,138& Passage  & \greencheck\\
QuAIL \cite{rogers2020getting} 
& MC-RC & Multi-domain & Multi  & Crowd  & 800 & Passage  & \greencheck\\
\hline
MovieQA \cite{tapaswi2016movieqa}     
& MM-QA   & Movie     & Movies     & Crowd  & 408 & Movie  & \emark \\
Visual7W \cite{zhu2016visual7w}  
& MM-QA   & Visual  & Images & Crowd & 47,300 & Image & \greencheck \\ 
TQA \cite{kembhavi2017you}  
& MM-QA & Science exam & Educational  & Expert  & 1,076 & Lesson  & \greencheck \\
RecipeQA \cite{yagcioglu-etal-2018-recipeqa} 
& MM-QA & Cooking  & Recipes & Auto & 19,779 & Recipe & \greencheck \\
ScienceQA \cite{lu2022learn} 
& MM-QA & Science exam  & Educational & Expert  &  21,208   & Query  & \greencheck \\
\hline
\end{tabular}
 }
 \vspace{-2mm}
\end{table*}

\subsection{Fill-in-the-Blank (FITB)} \label{sec:FITB}
\frenchspacing
Cloze queries, also known as fill-in-the-blank, are available in both textual \cite{xie-etal-2018-large} and visual \cite{yagcioglu-etal-2018-recipeqa} formats.
DGen dataset, illustrated in example \hyperlink{example1}{(1)}, presents a stem sentence with a placeholder and a set of options intended to fill that placeholder. 
The challenge is to create plausible yet incorrect distractors.

\vspace{-1mm}
\begin{quote}
\begin{small}
\hypertarget{example1}{(1)}
\textbf{Stem:} \textit{the organs of respiratory system are \_\_\\ }
\textbf{Distractors:} \textit{ a) ovaries,  b) intestines, c) kidneys}\\
\textbf{Answer:} \textit{lungs}
\vspace{-1mm}
\end{small}
\end{quote}

\subsection{Multiple-Choice Question (MCQ)} \label{sec:MC-DG}
\frenchspacing
For decades, research communities have shown interest in generating distractors for MCQ \cite{mitkov2003computer, bitew2022learning}. 
MCQ is divided into  (i) {\em question answering},  (ii) {\em reading comprehension}, and (iii) {\em multi-modal question answering}.

\vspace{1mm}
\noindent\textbf{Question Answering:}
A standard example of a multiple-choice question-answering task (MC-QA) is shown in example \hyperlink{example2}{(2)} from the SciQ dataset. 
The example presents a stem question with a set of options, including one correct answer and several in-context, yet incorrect distractors.

\vspace{-1mm}
\begin{quote}
\begin{footnotesize}
\hypertarget{example2}{(2)}
\textbf{Stem:} \textit{What eye part allows light to enter? \\ }
\textbf{Distractors:} \textit{a) iris,  b) retina, c) eyelid} \\
\textbf{Answer:} \textit{pupil}
\vspace{-1mm}
\end{footnotesize}
\end{quote}

\noindent\textbf{Reading Comprehension:} 
\frenchspacing
A typical example of a multiple-choice reading comprehension task (MC-RC) is displayed in example \hyperlink{example3}{(3)} from the RACE dataset.
The challenge involves generating distractors that are relevant to the given stem and passage, yet distinctly different from the answer. 

\vspace{-1mm}
\begin{quote}
\begin{footnotesize}
\hypertarget{example3}{(3)}
\textbf{Passage:} \textit{My name's Mary. This is my family tree ...
That boy is my brother. His name is Tony. This is Susan. She is my uncle's daughter. \\}
\textbf{Stem:} \textit{Tony and Mary are Susan's \_\_\_\_\_ \\ }
\textbf{Distractors:} \textit{ a) brothers, b) sisters, c) friends} \\
\textbf{Answer:} \textit{cousins}
\vspace{-1mm}
\end{footnotesize}
\end{quote}

\begin{figure}[H]
\begin{center}
 \includegraphics[width=0.8\textwidth, height=3.9cm] {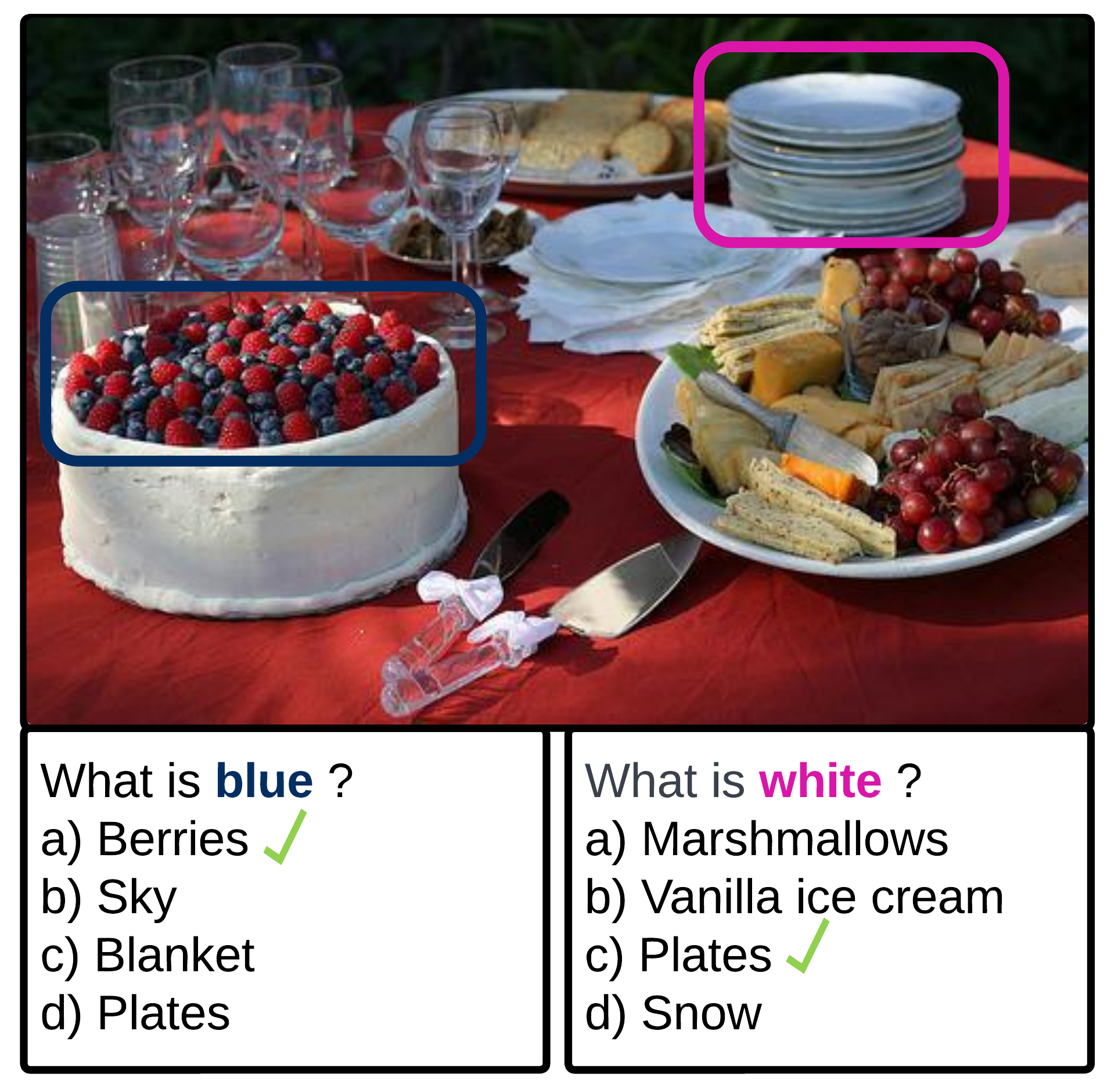}
    \caption{Multi-modal question answering task.}
    \label{fig:Visual7w}
    \vspace{-3mm}
\end{center}
\end{figure}

\noindent\textbf{Multi-modal Question Answering:} 
\frenchspacing
An example of a multi-modal question answering task (MM-QA) \cite{Lu_2022_CVPR}  is illustrated in Figure \ref{fig:Visual7w}. The distractors include all the options except for the correct answer, which is indicated by a green checkmark. The main challenge is to generate distractors that are relevant to the given question and image but are not correct as an answer.

\vspace{-1mm}
\section {Methods - Distractor Generation } \label{sec: Methods}
\vspace{-1mm}
\frenchspacing
The methods range from traditional to advanced AI approaches, including deep neural networks and pre-trained language models.

\subsection{Traditional Methods} \label{sec:traditional}
\frenchspacing
Traditional methods propose retrieving word-level distractors similar to an answer in specific domains.

{\em Corpus-based} methods rely on corpus features and syntactic rules in selecting distractors. 
\citet{chen-etal-2006-fast} used a part-of-speech tagger to transform an answer into various grammatical distractors, such as different verb tenses, in grammar cloze tests.  
\citet{pino2009semi} generated distractors through phonetic and morphological features. 
\citet{hill-simha-2016-automatic} utilized n-gram corpus to find potential distractors by filtering out all candidates that fit the context in cloze queries.
\citet{sakaguchi-etal-2013-discriminative} extracted distractors as error-correction pairs from a large English as a Second Language (ESL) corpus. 
\citet{agarwal-mannem-2011-automatic} followed part-of-speech similarity and term frequency to select distractors in biology cloze queries. 
\citet{zesch-melamud-2014-automatic} explored DG for verb cloze queries using context-sensitive inference rules \cite{melamud-etal-2013-two}, as it used the rules to filter out semantically similar distractors that are out of the context.
Corpus-based features are limited to simple distractors, often lacking plausibility in several domains as they fail to capture the semantic relationships required for contextually appropriate distractors.

% {\em Graph-based}
{\em Knowledge-based} methods retrieve distractors from hierarchical structures representing concepts and their relationships.
WordNet \cite{miller1995wordnet} and Probase \cite{wu2012probase} as knowledge-base examples are utilized to generate distractors in MC-QA \cite{mitkov2003computer, mitkov-etal-2009-semantic} and FITB \cite{pino2008selection}.
Notably, \citet{ren2021knowledge} proposed a framework using 
knowledge-base and contextual information from the question stem and key answer to construct a small set of semantically related distractors, which  
employs a probabilistic topic model
to determine the relevance of concepts to the key within the given stem. Knowledge-base contains static knowledge which may not 
be appropriate in specialized domains. 
Thus, an ontology-based method is utilized in distractor retrieving. \citet{stasaski-hearst-2017-multiple} used biology expert-curated concepts to select distractors that share some properties with the correct answer while differing in at least one key relationship to remain plausible but incorrect. 
\citet{leo2019ontology} utilized ontology in medical domain distractors.
\citet{kumar2023novel} utilized both knowledge-base and ontology as part of a generation system for collecting distractors in the technical education domain. 
Ontology, a static and domain-independent concept, may not cover all necessary concepts for diverse distractors. 
It is complex, time-consuming, and requires expert knowledge to ensure accuracy and relevance.

\vspace{-1mm}
\subsection {Deep Neural Network Models}\label{sec:Seq2Seq}
\frenchspacing
\vspace{-1mm}
Neural networks, including Sequence-to-Sequence (Seq2Seq)  \cite{NIPS2014_a14ac55a} models and attention mechanisms \cite{bahdanau2015neural}, showed success in DG
at word and sentence levels in MC-RC task.   
Seq2Seq models map input sequences such as passage, question, or answer to output sequence, a distractor, through conditional log-likelihood. MC-RC task handles long input sequence (e.g., a passage average token in RACE is 352.8) and requires distractors that are (i) semantically relevant to the passage, (ii) coherent with the question, and (iii) non-equivalent to the answer. 

Initially, \citet{gao2019generating} proposed a hierarchical encoder-decoder (HRED) network \cite{li-etal-2015-hierarchical} with two attention mechanisms. 
HRED showed superior performance in handling long input sequences tasks such as head-line generation \cite{ijcai2017p574} and summarization \cite{ling-rush-2017-coarse}. 
HRED encodes long given passages into word-level and sentence-level representations. 
A hierarchical dynamic attention allows both word-level and sentence-level attention distributions to change at each decoding time step to only focus on important sentences in the passage.
A static attention is proposed to learn the distribution of the sentences that are semantically relevant to the question rather than the answer.
In decoding, a special question-based initializer is used instead of encoder's last hidden state to generate a distractor that is grammatically consistent with the question. 

Several studies followed HRED network with other attention mechanisms. For example, \citet{zhou2020co} utilized co-attention mechanism \cite{seo2016bidirectional} to help the encoder better capture the rich interactions between the passage and question to generate relevant distractors. 
\citet{shuai2021topic} explored static attention with topic-enhanced multi-head co-attention through Latent Dirichlet Allocation (LDA) to calculate the topic-level attention between question and passage sentences. 
\citet{maurya2020learning} implemented the SoftSel operation \cite{tang2019multi} combined with a gated mechanism to eliminate answer-revealing sentences. 
Notably, \citet{shuai2023qdg} incorporate HRED into a question-distractor joint framework while other works mainly focused on DG task.

To generate multiple n-distractors, beam search with Jaccard distance is mainly utilized in several studies while \citet{maurya2020learning} explored multiple decoders.
\citet{xie2021diverse} proposed encoder-decoder multi-selector generation network (MSG-Net) based on mixture content selection \cite{cho-etal-2019-mixture} to generate diverse distractors based on n-sentence key selectors. 
The selected sentences are transformed into distractors using
T5 \cite{raffel2020exploring} as a generation layer.

% \begin{table}[h]
% \caption{Seq2Seq models in DG at MC-RC task. \textbf{HRED}: Hierarchical Encoder Decoder; \textbf{ED}: Encoder-Decoder}
% \label{tab:RC-DG Neural Models}
%  \resizebox{\textwidth}{!}{
% \begin{tabular}{lcc}
% \hline \hline
% Paper   & Model                        & Generation Method   \\ \hline
% HSA 
% (\citeyear{gao2019generating})    
% & HRED                                 & Beam Search \& Jaccard Distance  \\
% CHN 
% (\citeyear{zhou2020co})          
% & HRED                                 & Beam Search \& Jaccard Distance \\
% EDGE 
% (\citeyear{qiu-etal-2020-automatic})   
% & ED                              & Beam Search \& Jaccard Distance\\
% HMD-Net
% (\citeyear{maurya2020learning})
% & HRED(s)                              & Multiple Decoders \\
% TMCA
% (\citeyear{shuai2021topic})       
% & HRED                                 & Beam Search \& Jaccard Distance \\  
% MSG-Net 
% (\citeyear{xie2021diverse})   
% & ED                                   &  T5 model (\citeyear{raffel2020exploring}) \\
% QDG (\citeyear{shuai2023qdg})  & HRED  &      Beam Search \& Jaccard Distance \\  
% \hline   
% \end{tabular}
% \vspace{-5pt}
% }
% \end{table}

\subsection {Pre-trained Models} \label{sec:PLMs}
\frenchspacing
Pre-trained models, such as word2vec \cite{mikolov2013efficient}, GloVe \cite{pennington-etal-2014-glove}, and fastText \cite{bojanowski-etal-2017-enriching}, have revolutionized static word embedding generation. 
These models are commonly used in DG tasks like FITB \cite{kumar-etal-2015-revup, jiang-lee-2017-distractor, yoshimi-etal-2023-distractor} and MC-QA \cite{guo2016questimator} to select similar answer options using word vector cosine similarity. 
In the MC-RC task, \citet{susanti2018automatic} utilized word vector cosine similarity to select distractors for English vocabulary meaning.
% They are also utilized in MC-RC task \cite{susanti2018automatic}, specifically for selecting distractors for English vocabulary meaning using word vector cosine similarity. 

Pre-trained language models (PLMs) \cite{min2023recent} based on Transformer architecture \cite{vaswani2017attention} include \textbf{(i) auto-regressive} models such as GPT-models \cite{radford2019language, brown2020language}, \textbf{(ii) auto-encoding} models such as BERT \cite{devlin-etal-2019-bert}, and \textbf{(iii) encoder-decoder} (Text2Text) models such as T5 \cite{raffel2020exploring} and BART \cite{lewis-etal-2020-bart}. 
PLMs utilize {\em fine-tuning} and {\em prompting} methods in DG.

\begin{figure*}[!ht]
\begin{center}
    \includegraphics[width=1\textwidth]{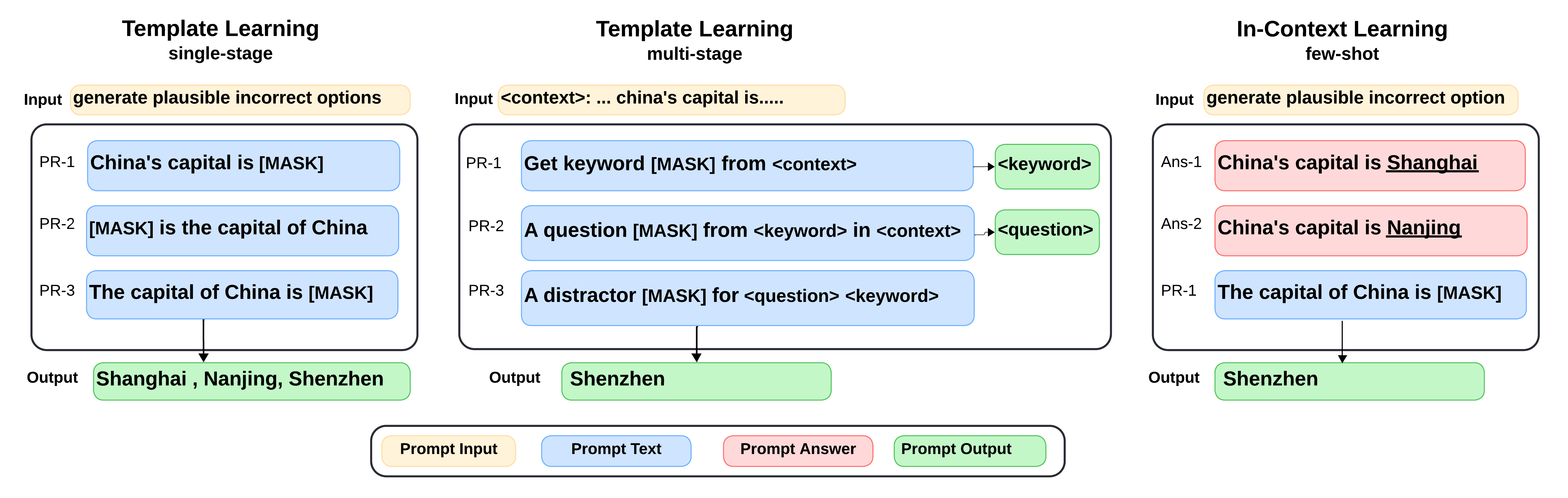}
    \caption{DG via prompting LLM. Figure is adapted from \cite{liu2023pre}.} 
    \label{fig:Prompting_methods}
    \vspace{-5mm}
\end{center}
\end{figure*}

\subsubsection{PLMs with Fine-Tuning} \label{sec:Fine-Tuning}
\frenchspacing
PLMs, pre-trained on large amounts of unlabelled data, can be fine-tuned on specific tasks using small labeled datasets. 
Table \ref{tab:Finetuning_DG} presents DG studies where PLMs with fine-tuning have been utilized.

In \textbf{auto-regressive} models, \citet{offerijns2020better} fine-tuned GPT-2 model trained on the RACE dataset to generate three distractors for a given question and context.

In \textbf{auto-encoding} models,  \citet{chung-etal-2020-bert} proposed BERT model as auto-regressive iterations with multi-tasking and negative answer regularization to generate distractors in MC-RC task.
\citet {chiang-etal-2022-cdgp} explored several PLMs instead of knowledge-base methods \cite{ren2021knowledge} for generating distractors in FITB task. 
The models are trained based on naive fine-tuning and answer-relating fine-tuning.
\citet{bitew2022learning} explored a multilingual BERT encoder to create context-aware neural networks in MC-QA. 
The model ranks distractors based on relevance to the question stem and answer key through contrastive learning.

In \textbf{Text2Text} models, \citet{wang-etal-2023-distractor} suggested T5 and BART models for FITB task. To boost model performance, candidate augmentation strategy and multi-tasking training techniques are utilized. 
% \textcolor{blue}{
\citet{yu-etal-2024-enhancing} applied a retrieval-augmented pre-training (RAP) approach and used knowledge graph triplet for data augmentation.
RAP method involves using answers to retrieve relevant sentences and passages  from a large corpus such as Wikipedia, masking these answers to create pseudo questions, and utilizing these questions to align T5 and BART models specifically for DG task.
% }
\citet{taslimipoor-etal-2024-distractor-generation} also proposed using T5 model for DG in MC-QA and MC-RC. 
The proposed approach utilized a two-step method: initially generating both correct and incorrect answers, and then discriminating between them with a classifier. 
The generated options are then clustered to remove duplicates and to ensure the diversity of the distractors.
T5 has been widely used in DG for MC-QA tasks 
related to questionnaires \cite{rodriguez2022end} and personalized exercises \cite{lelkes2021quiz, vachev2022leaf}.

\begin{table}[!h]
\centering
% \footnotesize
\caption{Fine-tuned PLMs on DG tasks.}
\label{tab:Finetuning_DG}
 \resizebox{\textwidth}{!}{
\begin{tabular}{llcc }
\hline \hline
Paper           & PLMS            & Language      &  Task      \\ \hline
{\scriptsize\cite{yeung-etal-2019-difficulty}}    & BERT (\citeyear{devlin-etal-2019-bert})       & Chinese       &  FITB \\ 
{\scriptsize\cite{chung-etal-2020-bert}}          & BERT (\citeyear{devlin-etal-2019-bert})       & English       &  MC-RC    \\
{\scriptsize\cite{offerijns2020better}}    & GPT-2 (\citeyear{radford2019language})      & English       &  MC-RC            \\ 
% {\scriptsize\cite{xie2021diverse}}         & T5 (\citeyear{raffel2020exploring})         & English       &  MC-RC            \\ 
{\scriptsize\cite{lelkes2021quiz}}         & T5 (\citeyear{raffel2020exploring})         & English       &  MC-QA              \\
{\scriptsize\cite{kalpakchi-boye-2021-bert}} & BERT (\citeyear{devlin-etal-2019-bert})       & Swedish       &  MC-RC            \\
{\scriptsize\cite{chiang-etal-2022-cdgp}}   & BERT (\citeyear{devlin-etal-2019-bert})       & English       &  FITB            \\
{\scriptsize\cite{chiang-etal-2022-cdgp}}   & SciBERT (\citeyear{beltagy-etal-2019-scibert})    & English       &  FITB            \\
{\scriptsize\cite{chiang-etal-2022-cdgp}}   & RoBERTa (\citeyear{liu2019roberta})    & English       &  FITB           \\
{\scriptsize\cite{chiang-etal-2022-cdgp}}   & BART (\citeyear{lewis-etal-2020-bart})         & English       &  FITB            \\
{\scriptsize\cite{vachev2022leaf}}   & T5 (\citeyear{raffel2020exploring})    & English       &  MC-QA  \\ 
{\scriptsize\cite{rodriguez2022end}} & T5 (\citeyear{raffel2020exploring})    & English       &  MC-QA  \\ 
{\scriptsize\cite{foucher2022word2course}} & T5 (\citeyear{raffel2020exploring})   & English  &  MC-QA            \\ 
{\scriptsize\cite{bitew2022learning}}      & mBERT (\citeyear{devlin-etal-2019-bert})      & Multi-lingual &  MC-QA            \\ 
{\scriptsize\cite{wang-etal-2023-distractor}}     & BART (\citeyear{lewis-etal-2020-bart})      & English       &  FITB            \\
{\scriptsize\cite{wang-etal-2023-distractor}}     & T5 (\citeyear{raffel2020exploring})         & English       &  FITB            \\
{\scriptsize\cite{hadifar2023eduqg}}       & T5 (\citeyear{raffel2020exploring})         & English       &  MC-QA            \\
{\scriptsize\cite{de2024distractor}}       & mT5 (\citeyear{raffel2020exploring})        & Spanish       &  MC-RC \\
{\scriptsize\cite{taslimipoor-etal-2024-distractor-generation}}       & T5 (\citeyear{raffel2020exploring})        & English       &  FITB \\
{\scriptsize\cite{taslimipoor-etal-2024-distractor-generation}}       & T5 (\citeyear{raffel2020exploring})        & English       &  MC-RC \\
{\scriptsize\cite{yu-etal-2024-enhancing}}       & T5 (\citeyear{raffel2020exploring})        & English       &  FITB \\
{\scriptsize\cite{yu-etal-2024-enhancing}}       &  BART (\citeyear{lewis-etal-2020-bart})       & English       &  FITB \\
\hline   
\end{tabular}
}
\vspace{-7pt}
\end{table}

\setlength{\textfloatsep}{4pt plus 0pt minus 6pt}

\subsubsection{PLMs with Prompting} \label{sec:Prompting}
\frenchspacing
Prompting \cite{liu2023pre} involves adding 
text to the input or output to encourage large language model (LLM) to perform specific tasks.
Figure \ref{fig:Prompting_methods} illustrates prompting-based learning methods.

{\em Template-based learning} uses multiple unanswered prompts at inference time to make predictions and has shown significant capabilities in generating distractors for FITB \cite{zu2023automated} and MC-QA \cite{doughty2024comparative} through single-stage prompting. 
\citet{maity2024novel} proposed multi-stage prompting, inspired by the chain of thought method \cite{wei2022chain}, to generate distractors for MC-QA based on a given text context.

{\em In-context learning} involves providing a few additional answered examples to demonstrate how the LLM should respond to the actual prompt. 
As shown in Table \ref{tab:Prompting},  in-context learning with zero and few-shot examples is also applied in MC-QA. 
In few-shot learning, examples are selected based on relevant questions retrieved by BERT-based ranking model \cite{bitew2022learning,bitew2023distractor}. Additionally, \citet{mcnichols2023exploring} explored k-nearest neighbor (KNN) examples for math distractor and feedback generation, and 
% \textcolor{blue}{
\citet{feng-etal-2024-exploring} asserted that KNN examples 
%outperforms 
outperform 
fine-tuning and chain-of-thought methods in math distractors.
% }

\begin{table*}[h]
\centering
% \footnotesize
\caption{Prompting large language models for DG tasks. LLM such as ChatGPT is selected based on OpenAI models such as (gpt-3.5-turbo), Codex (code-davinci-002) and GPT-3 (text-davinci-003) \cite{brown2020language}.}
\label{tab:Prompting}
 \resizebox{\textwidth}{!}{ 
\begin{tabular}{llccccc}
\hline \hline
Paper                        & LLM    & Method      & Prompting        & Language  & Domain &  Task    \\ \hline
{\scriptsize\cite{bitew2023distractor}}     & ChatGPT & In-Context & zero + few shots & Multi-lingual & Open-Domain &  MC-QA \\
{\scriptsize\cite{zu2023automated}}       & GPT-2   & Template   & single stage     & English       & Language proficiency &  FITB   \\
{\scriptsize\cite{tran2023generating}}    & GPT-3   & Template   & single stage     & English       & Programming &  MC-QA \\
{\scriptsize\cite{tran2023generating}}    & GPT-4   & Template   & single stage     & English       & Programming &  MC-QA \\
{\scriptsize\cite{mcnichols2023exploring}} & Codex   & In-Context & zero + few shots & English       & Math  & MC-QA         \\
{\scriptsize\cite{mcnichols2023exploring}}& ChatGPT & In-Context & zero + few shots & English       & Math  & MC-QA          \\
% {\scriptsize\cite{feng-etal-2024-exploring}} & \textcolor{blue}{ChatGPT} & In-Context & few shots & English & Math  & MC-QA        \\

{\scriptsize\cite{feng-etal-2024-exploring}} & GPT-4 & Template & multi-stage & English       & Math  & MC-QA     \\

{\scriptsize\cite{doughty2024comparative}} & GPT-4   & Template  & single stage   & English         & Programming  & MC-QA    \\
{\scriptsize\cite{maity2024novel}}        & GPT-4   & Template  & multi-stage    & Multi-lingual   & Open-Domain  & MC-QA     \\
{\scriptsize\cite{maity2024novel}}        & Codex   & Template  & multi-stage    & Multi-lingual   & Open-Domain  & MC-QA     \\
\hline   
\end{tabular}
}
\vspace{-5pt}
\end{table*}

\subsection{Other Models} \label{sec:Others}
\frenchspacing
{Other models proposed retrieving distractors from feature-based learning models for FITB \cite{ren2021knowledge} and MC-QA \cite{liang-etal-2018-distractor}. 
\citet{sinha2020ranking}  used a hybrid semantically aware neural network, consisting of a convolutional neural network and bidirectional LSTM, to retrieve distractors in an MC-QA task. 
These models have shown better performance compared to those using generative adversarial networks \cite{liang2017distractor}.
In domain-specific such as English Language test, round trip machine translation methods \cite{panda-etal-2022-automatic, palma-gomez-etal-2023-using} with alignment computation \cite{jalili-sabet-etal-2020-simalign} can generate a variety of distractors. 
In multi-modal, \citet{Lu_2022_CVPR} utilized reinforcement learning for textual DG, while \citet{ding-etal-2024-learn-question} proposed framework, using encoder-decoder vision-and-language model with contrastive learning to jointly generate questions, answers, and distractors.}

\section{Evaluation Methods} \label{sec:evaluation}
\frenchspacing
Evaluation methods for DG include {\em automatic} 
and {\em manual} approaches that rely on human judgment.

\subsection{Automatic Evaluation} \label{sec:automatic_E}
\frenchspacing
The automatic metrics are {\em ranking-based} \cite{valcarce2020assessing} and {\em n-gram} \cite{sai2022survey} metrics.

\subsubsection{Ranking-based Metrics} \label{sec:ranking}
\frenchspacing
Ranking-based metrics evaluate the model in retrieving relevant distractors across k-top locations.

{\em Order-unaware} metrics, which do not consider the order,  include Precision (P@K), Recall (R@K), and F1-score (F1@K). 
(P@K) calculates the ratio of correctly identified relevant distractors to the total number of options ranked within the top k positions. 
(R@K) measures the ratio of correctly identified relevant distractors to the total number of relevant distractors in the ground truth, and (F1@K) is the harmonic mean of precision and recall.
   
{\em Order-aware} metrics, which take the order into consideration, include Mean Reciprocal Rank (MRR@K), Normalized Discounted Cumulative Gain (NDCG@K), and Mean Average Precision (MAP@K). 
MRR@K focuses on the position of the first relevant item by averaging the reciprocal ranks of this item in the top k distractors across all queries.
NDCG@K compares the generated rankings to an ideal order, and 
MAP@K calculates the mean of average precision scores at k, considering the number and positions of relevant distractors. 
However, they struggle to identify semantic relatedness, multiple answers, or nonsensical distractors.

\subsubsection{N-gram Metrics} \label{sec:NLG}
\frenchspacing
N-gram metrics evaluate the word n-gram overlap between the hypothesis (i.e., generated distractors) and references (i.e., ground truth distractors). 
For example, BLUE \cite{papineni-etal-2002-bleu} is a precision-based metric calculating the ratio of n-grams between the hypothesis and references to the total n-grams in the hypothesis. 
Self-BLEU \cite{caccia2019language} measures lexical diversity between hypotheses. 
ROUGE \cite{lin-2004-rouge} is a recall-based metric calculating the ratio of n-grams between the hypothesis and references to the total n-grams in the reference. 
ROUGE-L uses F-score, where the precision and recall are computed to measure the longest common subsequence between sentence pairs.
METEOR \cite{lavie2009meteor} is an F-score metric that applies unigram matches, performing exact word mapping, stemmed word matching, and then synonym and paraphrase matching. Lexical mismatch may fail to identify valid distractors, leading to manual evaluation methods.

\subsection{Manual Evaluation}\label{sec:human_E}
\frenchspacing
The DG evaluation primarily relies on {\em plausibility} to ensure that distractors are semantically similar to the answer, grammatically correct within the query, and consistently relevant to the context, {\em reliability} to ensure incorrectness, and {\em diversity} to reflect the difficulty in identifying the correct answer. Thus, manual methods are utilized in this task.

{\em Comparative} method \cite{gao2019generating} selects the distractors based on specific objectives such as \textbf{confusion}, assessing the number of times a distractor being chosen as the best option without providing the correct answer, and \textbf{non-error} measuring the number of correct answers to a question. 

{\em Quantitative} method \cite{maurya2020learning} 
%relays 
relies on numerical scales within a specific range to evaluate a given objective. 
For instance, \textbf{reliability} and \textbf{plausibility} are the most essential metrics and participants use a 3-point scale for plausibility, and a binary mode for reliability for given generated and ground-truth distractors. 
Also, \textbf{fluency} assesses if a distractor follows proper language grammar, human logic, and common sense,  \textbf{coherence} evaluates distractor key phrases for relevance to the article and question, \textbf{distractibility} measures the likelihood of a candidate being chosen as a distractor, \textbf{diversity} measures semantic difference between multiple distractors, and \textbf{difference} measures the proportion of distractors and answer with the same semantics.
 
\section{Discussions and Findings} \label{sec:Findings}
\frenchspacing
This section provides analysis of the current AI models utilized for DG, along with an overview of the existing and emerging benchmarks. 

\subsection{Analysis of AI Models} \label{sec:analysis_AI_models}
\frenchspacing
\textbf{Do current models improve the quality of FITB and MC-QA tasks?} 
DG studies primarily focused on plausibility, but the reliability aspect has not been thoroughly studied. Static-based word embeddings like Word2Vec \cite{jiang-lee-2017-distractor} as shown in example (1) in Table \ref{tab:Incorrectness} are prone to generate multiple semantically correct answers, which fail to satisfy reliability. In contrast, dynamic context-based word embeddings like BERT \cite{devlin-etal-2019-bert} may produce compound names as distractors that are overly technical, which leads to the answer-revealing issue and fails to satisfy diversity. Feature-based learning models \cite{liang-etal-2018-distractor} might predict too easy options. PLMs are still susceptible to generating nonsense distractors, such as duplicate correct answers, obviously incorrect options, or previously generated distractors as shown in examples (2) and (3) in Table \ref{tab:Incorrectness}  through fine-tuning FITB task.  \citet{wang-etal-2023-distractor} utilized data augmentation to reduce these issues.
% \textcolor{blue}{
\citet{yu-etal-2024-enhancing} examined the use of knowledge graph triplets as a data augmentation technique during fine-tuning, noting that it might introduce noise that could interfere with the model generation process.
% }
Few-shot examples \cite{bitew2023distractor} reduced nonsense distractor rate in open-domain from 50\% to 16\%.
Thus, the quality of DG 
% in these tasks 
is still insufficient for reliable and diverse distractors. 

\begin{table}[h]
\centering
% \footnotesize
\caption{DG quality in FITB and MC-QA tasks.}
% \vspace{-20pt}
\label{tab:Incorrectness} 
 \resizebox{\textwidth}{!}{
\begin{tabular}{c|c|c}
\hline
\multicolumn{3}{l}{\begin{tabular}[c]{@{}l@{}}(1) \textbf{Stem} : The main source of energy in your body is --- \\ \hspace{6mm} \textbf{Answer}: \textcolor{blue}{carbohydrate}\end{tabular}} \\
\hline
Method           & Distractor  & Problem  \\ \hline    
EmbSim {\scriptsize(\citeyear{jiang-lee-2017-distractor})} & \textcolor{purple}{- glucose} & valid answer \\   
BERT {\scriptsize(\citeyear{devlin-etal-2019-bert})}  & \textcolor{purple}{- glycosaminoglycans} & too technical \\
LR+RF {\scriptsize(\citeyear{liang-etal-2018-distractor})}
& \textcolor{purple}{- methane}           & obviously wrong \\ \hline

\multicolumn{3}{l}{\begin{tabular}[c]{@{}l@{}} (2) \textbf{Stem}: Rural area do not have school, that is not ------- \\ 
\hspace{6mm} \textbf{Answer}: \textcolor{blue}{fair} \end{tabular}} \\ \hline
Method           & Distractor  & Problem  \\ \hline
T5 {\scriptsize(\citeyear{wang-etal-2023-distractor})} & 
\begin{tabular}[c]{@{}l@{}} 
\textcolor{purple}{- fair} 
\end{tabular} & similar to answer\\

BART {\scriptsize(\citeyear{wang-etal-2023-distractor})} &   
\textcolor{purple}{- unfair} 

& obviously wrong \\ \hline

\multicolumn{3}{l}{\begin{tabular}[c]{@{}l@{}}(3) \textbf{Stem}: She let people ----- more about Vietnam \\ 
\hspace{6mm} \textbf{Answer}: \textcolor{blue}{know}\end{tabular}} \\ \hline
Method           & Distractor  & Problem  \\ \hline
T5 {\scriptsize(\citeyear{wang-etal-2023-distractor})} & \begin{tabular}[c]{@{}l@{}} 
\textcolor{purple}{- think, think , think} 
% \\  \textcolor{purple}{- think}
\end{tabular}         
& previously generated \\ \hline
\end{tabular}

}
\end{table}

\begin{table}[h]
\vspace{-3mm}
\centering
% \footnotesize
\caption{DG validity in the MC-RC task. 
% Examples source is RACE dataset
}
\label{tab:Beam-Search}
 \resizebox{\textwidth}{!}{
\begin{tabular}{c|c}
\hline

\multicolumn{2}{l}{\begin{tabular}[c]{@{}l@{}}
(1) 
\textbf{Passage}: Nuclear power's danger to health ... etc\\
% \hyperlink{example12}{(12)} \\ 
\hspace{6mm}\textbf{Question}: Which of the following statements is true? \\ 
\hspace{6mm} \textbf{Answer}: \textcolor{blue}{Nuclear radiation can cause cancer in human beings}\\
\hspace{6mm} \textbf{Method}: HMD-Net \cite{maurya2020learning}
\end{tabular}} \\
\hline
Distractor  & Problem  \\ \hline
\begin{tabular}[c]{@{}l@{}} 
\textcolor{purple}{- Radiation is harmless,}\\
\textcolor{purple}{- Radiation can't hurt all over us,}\\
\textcolor{purple}{- Radiation can't kill human beings.}\\
\end{tabular}               
& 
\begin{tabular}[c]{@{}l@{}} 
lexically differ, but \\
semantically similar.
\end{tabular} 
\\ \hline

\multicolumn{2}{l}{\begin{tabular}[c]{@{}l@{}}
(2) 
\textbf{Passage}: Most of the time, people wear hats to protect ...etc\\
% given passage \hyperlink{example14}{(14)} \\ 
\hspace{6mm}\textbf{Question}: which of the women would look most attractive? \\ 
\hspace{6mm} \textbf{Answer}: \textcolor{blue}{A short red-haired woman who wears a purple hat} \hspace{6mm}\\
\hspace{6mm}\textbf{Method}: BDG \cite{chung-etal-2020-bert}
\end{tabular}} \\

\hline
 Distractor  & Problem  \\ \hline  
\begin{tabular}[c]{@{}l@{}} 
\textcolor{purple}{- young woman wears a white hat,}\\
\textcolor{purple}{- young woman wears a white hat,}\\
\textcolor{purple}{- short woman with big, round faces}.\\
\end{tabular}               
& 
\begin{tabular}[c]{@{}l@{}} 
previously generated\\ and biased options
% options
\end{tabular} 
\\ \hline

\multicolumn{2}{l}{\begin{tabular}[c]{@{}l@{}}
(3) 
\textbf{Passage}: About a third of all common cancers ...etc\\
% given passage \hyperlink{example13}{(13)} \\ 
\hspace{6mm}\textbf{Question}:  By writing the passage, the author mainly intends to \-\-\- \\ 
\hspace{6mm} \textbf{Answer}: \textcolor{blue}{Advice people to develop healthier lifestyle} \\
\hspace{6mm} \textbf{Method}: MSG-Net \cite{xie2021diverse}
\end{tabular}} \\

\hline
Distractor  & Problem  \\ \hline 
\begin{tabular}[c]{@{}l@{}} 
 \textcolor{purple}{- teach people how to prevent cancers,}\\
 \textcolor{purple}{- advice people to stop smoking,}\\
 \textcolor{purple}{- protect people from developing cancer.}\\
\end{tabular}               
& 
\begin{tabular}[c]{@{}l@{}} 
lack difficulty control
\end{tabular} 
\\ \hline

\end{tabular} 
}
% \vspace{-2pt}
\end{table} 

\setlength{\textfloatsep}{4pt plus 0pt minus 6pt}

\vspace{1mm}
\noindent\textbf{Are current models satisfied validity in MC-RC task?}
Despite the use of dynamic and static attentions 
in MC-RC models for plausibility and reliability, there are still shortcomings.
The beam search methods \cite{gao2019generating, shuai2023qdg} in Seq2Seq models fail to generate diverse distractors. Also, multi-decoders \cite{maurya2020learning} as demonstrated in example (1) in Table \ref{tab:Beam-Search}  used a mixture of decoders in decoding stage to generate divers distractors, but distractors are generated from the same input and have identical semantics which leads to options that are lexically diverse, but they are semantically similar. These generation methods cause an answer-revealing issue.
PLMs are still vulnerable to answer copying and biased options \cite{chung-etal-2020-bert},  as shown in example (2) in Table \ref{tab:Beam-Search}. The content selection approach \cite{xie2021diverse} in example (3) in Table \ref{tab:Beam-Search} can generate diverse distractors from different sentences, but further exploration or implicit common sense reasoning is required for difficult controls.
Thus, the validity of DG has room for improvement. Quantitative comparisons are  detailed for DG tasks in Appendix~\ref{sec:quantitative_results}, 
% \textcolor{blue}{
providing performance metrics and results for recent AI models utilized for DG tasks.
% }.

\subsection{Analysis of Benchmarks} \label{sec:analysis_benchmarks}
\frenchspacing

\vspace{1mm}
\noindent{\textbf{Are low-resource datasets explored in DG?}
Despite the use of English datasets, low-resource datasets remain limited in DG.
Pioneering research explored DG in
Spanish \cite{de2024distractor}, Swedish \cite{kalpakchi-boye-2021-bert}, Chinese \cite{yeung-etal-2019-difficulty}, Japanese \cite{andersson2023closing} and others \cite{maity2024novel} including German, Bengali, and Hindi. Typically, small-scale datasets or translated English datasets are used to create these training data.
Notably, there are efforts to build non-English multiple-choice datasets in French \cite{labrak-etal-2022-frenchmedmcqa}, Chinese \cite{sun-etal-2020-investigating}, Bulgarian \cite{hardalov-etal-2019-beyond}, Vietnamese \cite{van2020enhancing} and a multi-lingual \cite{bitew2022learning} datasets.
These datasets enable low-resource DG exploration and highlight the need for non-English datasets across various domains and tasks.
}

\vspace{1mm}
\noindent{\textbf{Are open-domain datasets emerging in DG?}
Specific domains such as Science (e.g., SciQ) or English (e.g., CLOTH) are utilized in DG, but there are limited open-domain datasets (e.g., Televic, EduQG) emerging in the field.
For example, Televic, which covers multiple subjects and includes multi-lingual content, contributes significantly to DG by posing new challenges, such as generating nonsensical distractors \cite{bitew2022learning, bitew2023distractor}.

\section{Future Directions}  \label{sec:Future_work}
This section outlines directions for future research.

\subsection{Trustworthy Generation}
\frenchspacing
AI advancements in DG are improving, but they still face challenges like hallucination issues in PLMs \cite{ji2023survey} and 
a heavy reliance on costly human-annotated labels \cite{qu-etal-2024-unsupervised}. To control this task generation \cite{zhang2023survey}, reinforcement learning from human feedback (RLHF) \cite{ouyang2022training} and few-shot examples \cite{bitew2023distractor} may be 
utilized to improve the trustworthiness of DG.
Integrating knowledge-based methods has been proposed \cite{yu-etal-2024-enhancing} and further improvements may enhance the performance of PLMs. Also, pioneering works can train models to distinguish between valid and invalid distractors through advanced learning approaches such as 
% \textcolor{blue}{
contrastive learning \cite{NEURIPS2022_0f5fcf4b} that enables models to differentiate between semantically similar and dissimilar data pairs in the embedding space. This method has 
%showed 
shown 
significant improvement in enhancing representation learning by encouraging models to capture semantic relationships.
As a result, it has demonstrated notable success across various NLP tasks, including machine translation \cite{pan-etal-2021-contrastive}, text classification \cite{chen2022contrastnet}, and question answering \cite{karpukhin-etal-2020-dense}. Additionally, incorporating adversarial learning approaches \cite{li-etal-2023-adversarial, zhuang-etal-2024-trainable} may enhance the robustness of DG models.
% }

\subsection{Deployment in Education}
\frenchspacing
Distractor quality is crucial in personalized learning \cite{vachev2022leaf,lelkes2021quiz, li2024generating}, but the task remains challenging with current existing approaches \cite{dutulescu2024beyond} and evaluating their effectiveness in education remains an open research problem.
AI models explored LLMs ability to generate multiple-choice questions  that meet course learning objectives in the programming domain \cite{doughty2024comparative} and in various formats \cite{tran2023generating}.
LLMs have shown promise in generating usable multiple-choice questions in different domains and tasks, but their alignment with Bloom's Taxonomy levels still has significant room for improvement \cite{hwang2024towards}. Controlling the difficulty levels of generated candidates continues to be a major challenge for the NLP community, highlighting the necessity for additional research to create usable DG models.
Thus, instructors in education must ensure the quality of automated DG models by verifying its plausibility, reliability, diversity, alignment with learning objectives, and adherence to ethical guidelines. 

\subsection{Multi-Modal Generation}
\frenchspacing
The novel task \cite{Lu_2022_CVPR}, textual DG in visual question answering, faces two potential challenges.
First, there are potential needs in generating distractors for various multi-modal domains as recent studies \cite{ding-etal-2024-learn-question} mainly used Visual7w as a visual question answering dataset.
Multi-modal supported content, such as figures \cite{wang2021results}, charts \cite{kafle2018dvqa}, and tables \cite{lu2023dynamic}, are available and used in different domains, including science \cite{kembhavi2017you} and mathematics \cite{verschaffel2020word} such as math word problem \cite{lu2021iconqa} and geometry problem solving \cite{chen-etal-2021-geoqa,lu-etal-2021-inter, chen-etal-2022-unigeo}.  
Second, research should focus on visual DG, specifically images, and incorporate videos and audios for new insights. These multi-modal insights could lead to novel applications and challenges in visual DG.

\subsection {Quality Metrics}
\frenchspacing
Current automatic metrics (e.g., n-gram) showed significant limitations such as excluding acceptable candidates due to lexical mismatching. 
Although some metrics can perform synonym n-gram matching (e.g., greedy matching \cite{rus-lintean-2012-comparison}, embedding average metrics \cite{john2016towards}, and vector extrema \cite{forgues2014bootstrapping}),
they cannot determine if semantic similarity will cause reliability issues such as multiple-answer problems.
Self-BLEU cannot ensure diversity, as it measures diversity in terms of lexical differences, which does not guarantee the difficulty of the distractors. 
Thus, few studies \cite{moon-etal-2022-evaluating, raina-etal-2023-assessing}  proposed systems for the quality of DG even though generalizing quality metrics in DG is still challenging.
Also, the assessing for nonsense distractors in open-domain \cite{bitew2022learning} still relies on manual metrics such as nonsense distractor rate.
Notably, item-writing flaws (IWFs) rubric evaluates the pedagogical value of both questions and options, serving as an essential quality evaluation tool in education. Ongoing research aims to automate this rubric \cite{moore2023assessing}, leading to advancements in automated quality assessment.

\section {Conclusion} \label{sec:conclusion}
\frenchspacing
Distractor Generation (DG) is critical in assessment and has received significant attention with advanced AI models. 
This paper surveys DG  tasks, including  fill-in-the-blank and multiple-choice question across text and multi-modal domains.
% \textcolor{blue}{
It categorizes the tasks within relevant datasets and provides a comprehensive analysis of the components in the available datasets.
% }
This paper also provides a detailed discussion of the current methods, summarizes the evaluation metrics, and discusses the main findings, including the analysis of AI models and benchmarks. It also outlines potential future research directions to facilitate further improvements and explorations. To enhance research in distractor generation, this paper also provides a continuously updated reading list available on a GitHub repository at \url{https://github.com/Distractor-Generation/DG_Survey}.

% benchmarks, evaluation methods, and potential future research directions. 
% and a continuous updated reading list\footnote{\url{https://github.com/Distractor-Generation/DG_Survey}.}.

% Distractor Generation (DG) is critical in assessment and has received significant attention with advanced AI models. 
% This paper surveys DG  tasks, including  fill-in-the-blank and multiple-choice question across text and multi-modal domains.
% It categorizes DG tasks within relevant datasets and discusses the associated methods and evaluation metrics. 
% This paper also provides a detailed discussion of current methods, benchmarks, potential future research directions, \textcolor{blue}{and continuous updated reading list for DG at }{\url{https://github.com/Distractor-Generation/DG_Survey}}.

% \newpage 
\section*{Limitations}
\frenchspacing
This survey paper focuses on contemporary research in distractor generation problem using advanced AI methods, but it may not cover the entire historical scope and recent advancements that have emerged around the time or after the survey was conducted due to rapid research development.
% \textcolor{blue}{
Furthermore, the evaluation of existing models and benchmarks relies on recently collected papers and may not fully represent the %of 
state-of-the-art models for distractor generation tasks. 
% }
However, our survey is the first to comprehensively address distractor generation tasks and methods, providing detailed outlines of current datasets and evaluation methods.
It also provides a concise overview of the main findings, challenges, and potential future research directions, making it a valuable resource for scholars in the field.

\section*{Acknowledgments}
\frenchspacing
We would like to express our sincere gratitude to the anonymous reviewers for their constructive feedback and invaluable recommendations. This work was conducted with the collaborative support of Macquarie University and the University of Adelaide in Australia, along with Umm Al-Qura University in the Kingdom of Saudi Arabia. Furthermore, we are thankful for the support and encouragement received from the members of the Intelligent Computing Laboratory in the School of Computing at Macquarie University.

% We extend our sincere gratitude to the anonymous reviewers for their valuable feedback and insightful suggestions. This work was collaboratively supported by the School of Computing at Macquarie University, the School of Computer Science at the University of Adelaide, and the College of Engineering and Computing at Umm Al-Qura University. We would also like to thank the members of the Intelligent Computing Laboratory (ICLers) at Macquarie University for their continuous support.

\bibliography{anthology,custom}
\bibliographystyle{acl_natbib}
\appendix

% Appendix ----------------------------

\section {Multiple Choice Components} \label{sec:dataset_analysis}  
\frenchspacing
The fundamental components of a multiple-choice data item consist of (i) a {\em stem}, the query or question, (ii) an {\em answer}, the only true option, and (iii) a set of {\em distractors}, the set of false options. 
A {\em supported content} can be a given text, an image, or a video.

\subsection{Stem} \label{sec:stem}
\frenchspacing
A stem can be formed as a complete declarative sentence, a declarative sentence or passage with placeholders, a factoid query such as a deep level (why? how?) or shallow level (who? where?) in Bloom's taxonomy, or other non-factoid queries.  
It can also be formed as an image or a video in a multi-modal domain. 

\vspace{1mm}
\noindent\textbf{Fill-in-the-Blank (FITB)}: selecting an appropriate word, sentence, or an image to complete a given content or a query is known as cloze or FITB.
In textual data, CLOTH \cite{xie-etal-2018-large} in example \hyperlink{example4}{(4)} describes stem passage, and DGen \cite{ren2021knowledge} in example \hyperlink{example5}{(5)} indicates stem sentence while RecipeQA \cite{yagcioglu-etal-2018-recipeqa} data in Figure \ref{fig:visual_stem} outlines a visual stem.

%CLOTH
\vspace{-2mm}
\begin{quote}
\begin{footnotesize}
\hypertarget{example4}{(4)}
\textbf{Stem:} \textit{Nancy had just got a job as a secretary in a company. Monday was the first day she went to work, so she was very -- 1 -- and arrived early. She -- 2 -- the door open and found nobody ...}\\    
\textbf{Distractors -1-:} \textit{a) depressed, b) encouraged, c) surprised}\\
\textbf{Distractors -2-:} \textit{a) turned, b) knocked, c) forced }\\
\textbf{Answer -1- :} \textit{excited}\\
\textbf{Answer -2- :} \textit{pushed}
\end{footnotesize}
 \end{quote}
 
%DGEN
\begin{quote}
\begin{footnotesize}
\hypertarget{example5}{(5)}
\textbf{Stem:} \textit{the organs of respiratory system are \_\\ }
\textbf{Distractors:} \textit{a) ovaries, b) intestines, c) kidneys}\\
\textbf{Answer:} \textit{lungs}
\end{footnotesize}
\end{quote}

\begin{figure}[h]
\begin{center}
 \includegraphics[width=0.9\textwidth, height=6cm] {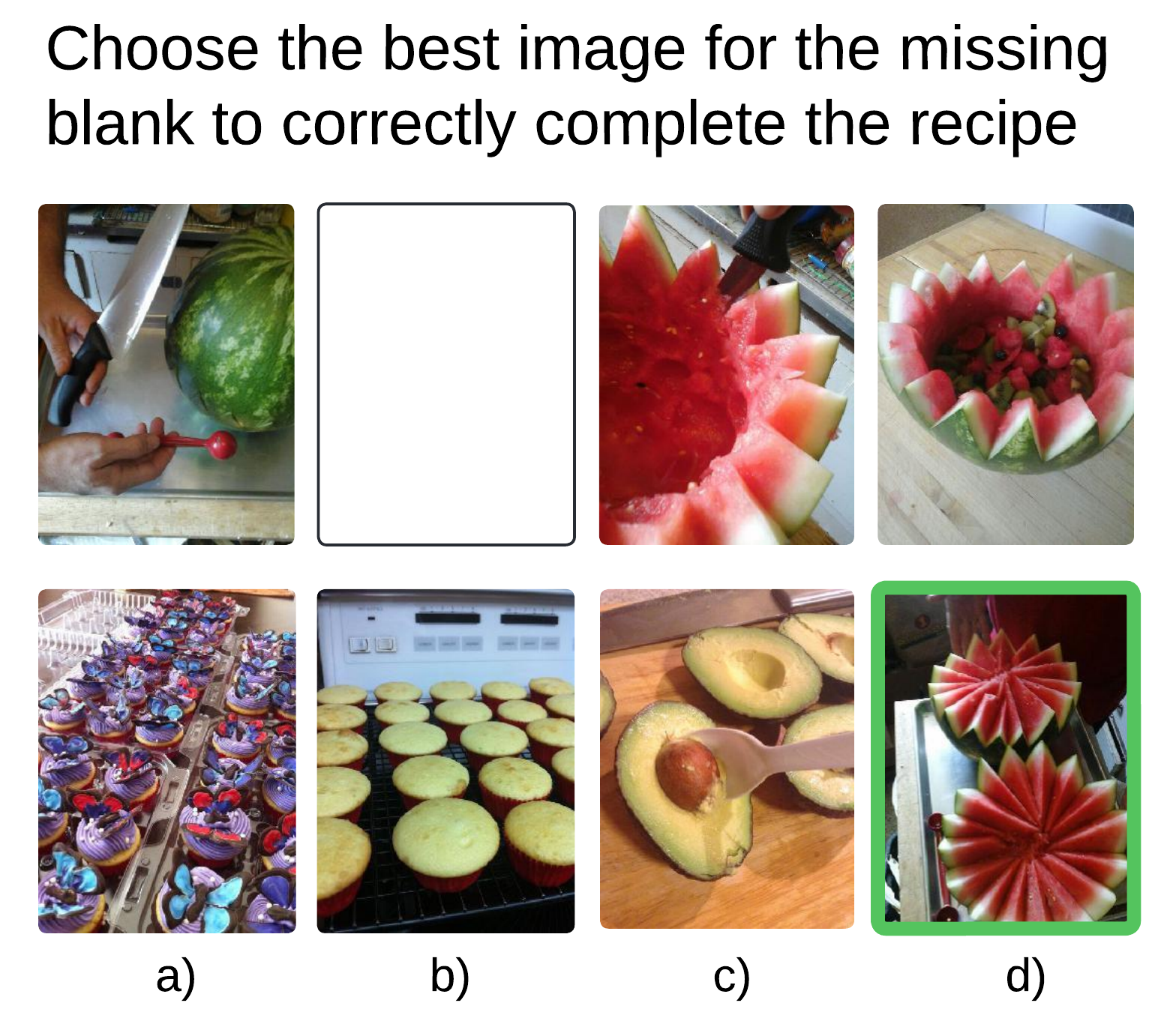}
    \caption{Visual Cloze. }
    \label{fig:visual_stem}
    \vspace{-5mm}
\end{center}
\end{figure}

\noindent\textbf{Multiple-Choice Question (MCQ)}: forming a question as a Wh-Q or declarative sentence is common in the MC-QA task. 
SciQ \cite{welbl-etal-2017-crowdsourcing} data in example \hyperlink{example6}{(6)} and  MCQL \cite{liang-etal-2018-distractor} data in example \hyperlink{example7}{(7)} illustrate textual factoid and declarative sentence stems, respectively.

%SciQ
\begin{quote}
\begin{footnotesize}
\hypertarget{example6}{(6)}
\textbf{Passage:} \textit{All radioactive decay is dangerous to living things, but \underline{alpha decay} is the least dangerous.}\\ 
\textbf{Stem:} \textit{What is the least dangerous radioactive decay?}\\  
\textbf{Distractors:} \textit{a) zeta decay, b) beta decay, c) gamma decay}\\
\textbf{Answer:} \textit{alpha decay}
\end{footnotesize}
\end{quote}

%MCQL
\begin{quote}
\begin{footnotesize}
\hypertarget{example7}{(7)}
\textbf{Stem:} \textit{During dark reactions, energy is stored in molecules of}\\    
\textbf{Distractors:} \textit{a) carbon, b) oxygen, c) hydrogen }\\
\textbf{Answer:} \textit{sugar}
\end{footnotesize}
\end{quote}

\subsection{Answer} \label{sec:answer}
\frenchspacing
An answer, also known as the correct option, must be unique for each query.
It can be formed as a textual short phrase or a sentence. 
It can also be extractive from a given passage or free-form generated from a supported passage or prior knowledge. 
It can also be an image as indicated in Figure \ref{fig:visual_stem}.

\vspace{1mm}
\noindent\textbf{Short or Long Phrase}: MCQL data in example \hyperlink{example7}{(7)} describes word-level answer, while RACE \cite{lai-etal-2017-race} data in example \hyperlink{example8}{(8)} demonstrates a long-sentence answer.

%RACE
\begin{quote}
\begin{footnotesize}
\hypertarget{example8}{(8)}
\textbf{Passage:} \textit{Homework can put you in a bad mood ... Researchers from the University of Plymouth in England doubted whether mood might affect the way kids learn ...} \\
\textbf{Stem:} \textit{Researchers did experiments on kids in order to find out \_\_\_ .}\\  
\textbf{Distractors:} \textit{a) how they really feel when they are learning, b) what methods are easy for kids to learn, c) the relationship between sadness and happiness}\\
\textbf{Answer:} \textit{whether mood affects their learning ability}
\end{footnotesize}
\end{quote}

\noindent\textbf{Extractive or Free-Form}: SciQ in example \hyperlink{example6}{(6)} describes an extractive answer type, where the answer is a span from the supported content, while MCQL in example \hyperlink{example7}{(7)} features a free-form answer type.

\subsection{Option} \label{sec:option}
\frenchspacing
All options, also known as distractors or false candidates, must be incorrect candidates to satisfy objectivity. 
Similar to the answer, options may 
be formed as words or sentences, mostly separated with each query but SCDE \cite{kong-etal-2020-scde} introduced shared options across all queries. Figure \ref{fig:visual_stem} shows visual options where (d) is the correct answer and others are image distractors.

\vspace{1mm}
\noindent\textbf{Separated or Shared}: CLOTH in  example \hyperlink{example4}{(4)} describes separated options, while SCDE in  example \hyperlink{example9}{(9)} shows shared options.

\begin{quote}
\begin{footnotesize}
\hypertarget{example9}{(9)}
\textbf{Stem:} \textit{-- 1 -- Now it becomes popular and people are dyeing their hair to make it different. Dyeing hair ... Since the base of hair is the scalp, you may have an allergic reaction. -- 2 -- You can follow them even when you are applying dye to your hair at home. -- 3 -- ...} \\        
\textbf{Shared Distractors:} \textit{(A) Colorful hair speaks more about beauty, (B) While dyeing your hair it is important to take some safety measures, (C) Don't forget to treat grandparents with respect because they're an essential part of your family, (D) It is better to apply hair dye for a few minutes... }\\
\textbf{Answers:} \textit{(1-A) (2-B) (3-D)...}
\end{footnotesize}
\end{quote}

\subsection{Supported Content} \label{sec:supported_content}
\frenchspacing
Supported content can take either a textual form (e.g., sentence, passage, or any form of text) or a visual form (e.g., image or video). 
Textual-supported content such as passage in the reading comprehension task is essential for assessing the examinee in real assessment. 
However, supported text content in datasets like SciQ is not primarily provided for reading comprehension tasks, while AQUA-RAT \cite{ling-etal-2017-program} provides rationales (i.e., mathematical equation formats) to create mathematical multiple-choice datasets.
Table \ref{tab:Multiple Choice Datasets} presents the classification of collected datasets in DG tasks.

\vspace{1mm}
\noindent\textbf{Textual Form}: OpenBookQA \cite{mihaylov-etal-2018-suit} in \hyperlink{example10}{(10)} describes supported sentence text while RACE \cite{lai-etal-2017-race} in example \hyperlink{example8}{(8)} describes passage content.
%OpenBookQA
\begin{quote}
\begin{footnotesize}
\hypertarget{example10}{(10)}
\textbf{Sentence:} \textit{the sun is the source of energy for physical cycles on Earth} \\
\textbf{Stem:}  \textit{The sun is responsible for}\\
\textbf{Distractors:} \textit{a) puppies learning new tricks, b) children growing up and getting old, c) flowers wilting in a vase  }\\
\textbf{Answer:} \textit{plants sprouting, blooming and wilting}
\end{footnotesize}
\end{quote}

\vspace{1mm}
\noindent\textbf{Visual Form}: Visual7W data in Figure \ref{fig:Visual7w} shows an image as supported content, while MovieQA \cite{tapaswi2016movieqa} data uses a movie as supported content.

% Datasets Statistics
\begin{table*}[tb!]
\caption{Dataset analysis of multiple-choice components. \xmark : not available, * : available upon request.}
\label{tab:Multiple Choice Analysis}
\resizebox{\textwidth}{!}{
\begin{tabular}{lccccccccccc}
\hline \hline
Dataset & Supported Content & Most Query Type & \#Passage ($P$) & \#Query ($Q$) & \#Option ($O$) & $P_{avg}$ & $Q_{avg}$ & $O_{avg}$ & $P_{vcb}$ & $Q_{vcb}$ & $O_{vcb}$ \\ \hline
CLOTH  
 & \xmark & Passage-Blank  & 7,131  & 99,433 & 4  & 329.8 & \xmark &  1 &  22,360 & \xmark & 7,455    \\

CLOTH-M 
 & \xmark & Passage-Blank  & 3,031 &  28,527 & 4 & 246.3 & \xmark & 1 &9,478 & \xmark & 3,330    \\

CLOTH-H 
 & \xmark  & Passage-Blank  & 4,100  & 70,906 & 4 & 391.5 & \xmark & 1 &19,428  & \xmark & 6,922    \\
 
SCDE      
 & \xmark  & Passage-Blank   & 5,959   & 29,731 & 7 & 248.6 & \xmark & 13.3 & 21,410 & \xmark & 12,693    \\
 
DGen
 & \xmark  & Sentence-Blank  &  \xmark &  2,880  & 4 & \xmark & 19.5 & 1 & \xmark & 4,527 & 3,630   \\ 
 
CELA   
 & \xmark  & Passage-Blank  & 150 &  3,000 & 4 & 408.5  & \xmark  &  1.3 & 3,500 & \xmark & 3,716    \\
\hline

SciQ 
 & Text  & Question  & 12,252  & 13,679 & 4  & 78 & 14.5 & 1.5 & 20,409 & 7,615 & 9,499   \\
 
AQUA-RAT 
 & Text &  Question & 97,975  &   97,975  & 5   & 52.7 & 37.2 & 1.6  &127,404  &31,406  & 76,115  \\
 
OpenBookQA 
 & Text &  Sentence & 1,326 &   5,957 & 4  & 9.4 & 11.5 & 2.9 & 1,416 & 4,295 & 6,989 \\
 
ARC  
 & \xmark  & Question  & \xmark &  7,787 & 4 & \xmark  & 22.5 & 4.6 & \xmark & 6,079 & 6,164    \\
 
ARC-Challange 
 & \xmark & Question  & \xmark & 2590 & 4 & \xmark & 24.7 & 5.5 & \xmark & 4,057 & 4,245 \\
 
ARC-Easy  
 & \xmark & Question   & \xmark  & 5197  & 4 & \xmark & 21.4 & 4.1 & \xmark & 4,998 & 5,021   \\
 
MCQL 
 & \xmark & Sentence  & \xmark &  7,116 & 4 & \xmark & 9.4 & 1.2 & \xmark & 5,703 & 7,108   \\
 
CommonSenseQA  
 & \xmark  & Question   & \xmark & 12,102 & 5  & \xmark & 15.1& 1.5 & \xmark & 6,844 &  6,921  \\
 
MathQA 
 & Text & Question  & 37,297  &  37,297  & 5   & 63.3  &  38.2  &  1.7 & 16,324  & 10,629  & 11,573    \\
 
QASC 
 & \xmark & Question   & \xmark  &  9,980  & 8 & \xmark &9.1 & 1.7 & \xmark & 3,886 & 6,407    \\
 
MedMCQA
 & Text  & Sentence  & 163,075  & 193,155  & 4 & 92.7  &  14.3   & 2.8  & 370,658 & 53,010 & 65,773     \\
 
Televic 
 & \xmark &  * & \xmark & 62,858 & $>$2  & \xmark & * &  * & \xmark & * & *    \\
 
EduQG 
& Text  & Multi-Form  & 3,397 &  3,397 & 4  & 209.3 & 16.3 & 4.2 & 21,077 & 5,311 & 8,632 \\
 
\hline

ChildrenBookTest
 & Text & Sentence-Blank  & 687,343 &  687,343 & 10  & 474.2  & 31.6 & 1 & 34,611 & 32,912 & 23,253  \\
 
Who Did What
 & Text & Sentence-Blank  & * & 205,978 & 2..5 &  *  &31.4 & 2.1 & * & 70,198 & 82,397  \\
 
MCTest-160      
 & Text  & Question  & 160 &  640 & 4 & 241.8 & 9.2 & 3.7 & 1,991 & 802 & 1,481   \\
 
MCTest-500      
 & Text  & Question  & 500 &  2,000 & 4  &  251.6  &  8.9 & 3.8 &3,079 & 1,436  & 23,34   \\
 
RACE     
 & Text & Sentence-Blank   & 27,933 & 97,687 & 4 & 352.8 & 12.3 & 6.7 & 88,851 & 20,179 & 32,899 \\
 
RACE-M 
 & Text  & Sentence-Blank  & 7,139  & 28,293 & 4  & 236  & 11.1 & 5  & 21,566 & 6,929 & 11,379 \\
 
RACE-H
 & Text & Sentence-Blank  & 20,784 & 69,394 & 4 & 361.9  & 12.4 & 6.9 & 81,887 & 18,318 & 29,491  \\
 
RACE-C 
 & Text & Sentence-Blank  & 4,275  &  14,122 & 4 & 424.1  & 13.8 & 7.4 &34,165 & 10,196 & 15,144  \\
 
DREAM  
 & Text & Question  & 6,444  & 10,197 & 3 & 86.4 & 8.8 & 5.3 & 8,449 & 2,791 & 5,864  \\
 
CosmosQA 
 & Text  & Question   & 21,866 & 35,588 & 4 & 70.4 & 10.6 &  8.1  & 36,970 & 10,685 & 18,173    \\
 
ReClor 
 & Text &  Question & 6,138 &   6,138 & 4 & 75.1 & 17 & 20.8 & 15,095 & 3,370 & 13,592   \\
 
QuAIL 
 & Text  & Question  &  800 &  12,966 & 4  & 395.4 & 9.7 & 4.4 & 13,750 & 6,341 & 9,955 \\
 
\hline

MovieQA    
 & Text + Video  & Question  & * & 14,944 & 5 & * & 10.7  &  5.6    & * & 7,440 & 15,242   \\
 
Visual7W  
 & Image &  Question & \xmark  &   327,939 & 4 &\xmark& 8 & 2.9   & \xmark & 12,168 & 15,430   \\
 
TQA 
 & Text + Image & Question   & 1,076  & 26,260 & 2..7 &  241.1  & 10.5 & 2.3  & 8,304 & 7,204 & 9,265  \\
 
RecipeQA 
 & Text + Image  & Sentence-Blank   &  19,779   & 36,786 & 4 & 575.1  & 10.8   & 5.7 & 78,089 & 5,587 & 71,369   \\
 
ScienceQA  
 & Text + Image & Question  &  10,220  &  21,208   &  >2  &  41.3 & 14.2   & 4.9   & 6,233  & 7,373  & 7,638  \\
\hline           
\end{tabular}
}
\end{table*}

\section{Multiple-Choice Datasets} \label{sec:analysis}
\frenchspacing
We collected multiple-choice datasets, as shown in Table \ref{tab:Multiple Choice Datasets} for DG tasks. 
We also summarized dataset properties, including related domain, source of data, generation method, corpus size, and unit.
Table \ref{tab:Multiple Choice Analysis} presents an analysis of multiple-choice components, including average token, vocabulary size, and most frequent type of query.  

\subsection{Dataset Analysis}
\frenchspacing
We utilized dataset analysis as proposed by \citet{dzendzik-etal-2021-english} to process our heuristic rules and statistics.
Using spaCy\footnote{\url{https://spacy.io/}.}
tokenizer we determined the average token length and vocabulary size of queries, passages, and options. 
We determine the most common query type for each dataset, using our proposed 
heuristic rules\footnote{\url{https://github.com/Distractor-Generation/DG_Survey}}.

\subsubsection{Data Domains}
\frenchspacing
In our collection, 10 of 36 datasets are from English exam sources and 9 from Science exam sources.
ReClor is for standardized tests and 4 datasets (i.e., DGen, EduQG, QuAIL, Televic) are for multi-domain fields. 
One dataset from the medicine domain and 2 datasets focus on math word problems. 
Three datasets are designed for children stories, two datasets for narratives, and one dataset for news.
Three multi-modal datasets are domain-specific such as movie, visual answering, and cooking.

\subsubsection{Data Creation} 
\frenchspacing
30 out of 36 datasets are created by humans.
% For these 30 datasets,
18 of them are created by experts and 12 are created by crowd workers. Some datasets are web-crawled such as MCQL and others (i.e., CBT, WDW, RecipeQA, DGen, CELA) are auto-generated.

\subsubsection{Data Corpus}  
\frenchspacing
 The corpuses of 31 datasets are text-based and 5 are multi-modal. 
 15 out of 36 corpuses are passages, also known as story, narratives, and dialogue. 
 %5 
 Five datasets are based on sentence units, 
 %2 
 two datasets have math word problems, and 
 %3 
 three 
 datasets are based on queries.
 Five datasets corpuses are books, chapters, or medical topics, and two datasets are based on WorldTree facts. 
 One dataset is based on the CONCEPTNET triplet (i.e., knowledge graph with commonsense relationships).

\subsubsection{Data Sources}
\frenchspacing
Out of 36 datasets, 22 are from educational materials and 14 are from blogs, stories, movies, images, or recipe sources. 

\vspace{1mm}
    \noindent\textbf{Educational Resources}: CLOTH, SCDE, RACE, RACE-C, DREAM are collected from educational public websites in China. 
    SciQ is extracted from 28 textbooks.
    TQA and ScienceQA are collected from CK-12 foundation website and school science curricula, respectively. 
    MCQL and AQUA-RAT are Web-crawled. OpenBookQA is derived from WorldTree corpus \cite{jansen-etal-2018-worldtree}.
    QASC has 17 million sentences from WorldTree and CK-12. 
    ReClor is generated from open websites and books. 
    EduQG, Televic, and MedMCQA are collected from the Openstax website, Televic education platform, and medical exam sources, respectively.

    \vspace{1mm}
    \noindent \textbf{{Multi-Sources}}:  QuAIL is collected from fiction, news, blogs, and user stories. 
    DGen contents are from SciQ, MCQL, and other websites. 
    CELA is constructed from CLOTH dataset and four auto-generated techniques (i.e., randomized, one feature -part of speech POS \cite{hill2016goldilocks}, several features - POS, word frequency, spelling similarity \cite{jiang-etal-2020-know}, and neural round trip translation \cite{panda-etal-2022-automatic}).
    
\vspace{1mm}
    \noindent\textbf{Other Sources}: CBT is built based on Project Gutenberg books, MCTest is crowd sourced, and CommonSenseQA used CONCEPTNET \cite{speer2017conceptnet}.
    CosmosQA uses personal narratives \cite{gordon2009identifying} from the Spinn3r Blog Dataset \cite{burton2009icwsm} and crowd-sourcing to promote commonsense reasoning \cite{sap2019atomic}.
    MovieQA, Visual7W, and RecipeQA are built utilizing 408 movies, COCO images \cite{lin2014microsoft}, and cooking websites, respectively.

\subsubsection{Data Components} 
\frenchspacing
The only dataset introduced as multi-format by labeling and forming a query as cloze and normal is EduQG. 
Therefore, we used heuristic rules to find the most common query type (i.e., blank, sentence, or question).
The average token length and vocabulary size of passages, queries, and options are presented in Table \ref{tab:Multiple Choice Analysis}. 
We outline the following: 

\vspace{1mm}
    \noindent
     \textbf{Supported Content}: all datasets contain text-supported content except DGen, ARC, CommonSenseQA, MCQL, QASC, and Televic. 
     In multi-modal, some datasets such as RecipeQA and TQA contain text and images.
     Other datasets such as MovieQA contain movies and (Visual7W, ScienceQA) contain images.   
     
\vspace{1mm}
    \noindent\textbf{Query Size}: CLOTH has the largest number of questions among the FITB datasets. 
    In MCQ datasets, the largest number of science questions found in SciQ (14K) and in math dataset is AQUA-RAT (98K). 
    Televic contains (63K) questions, covering open-domain multi-lingual dataset\footnote{50\% Dutch then French and English comes next.}.
    Only 198 questions ($Q_{avg}$14.9, $O_{avg}$ 1.9 average token) are provided in the GitHub sample. 
    The most usable dataset in the comprehension task is RACE (98K). Visual7W (327.9K) presents the largest number of questions in multi-model. 

    \vspace{1mm}
    \noindent \textbf{Number of Options}: most datasets have 4 to 5 separated options, but the SCDE average is 7 shared options. 
    QASC contains 8 choices. 
    Televic and ScienceQA start with 2 choices.
    CBT has 10, DREAM contains 3, and TQA is ranged between 2 to 7.
    % options.

    \vspace{1mm}
    \noindent \textbf{Component Average Length}: queries range from 8.8 to 19.5, and passages from 9.4 to 408 tokens. 
    Word-to-phrase token options have 1 to 4, while sentence-long options have more than 4 tokens.
    ReClor has the longest option tokens (20.8).

    \vspace{1mm}
    \noindent \textbf{Component Vocabulary Size}: The vocabulary for passages ranges from 1.4K to 371K based on the number of unique lowercase token lemmas. 
    The vocabulary for the queries spans from 802 to 70.2K, and the options span from 1.5K to 82.4K.

\subsubsection{Data Usability and Availability}
\frenchspacing
 Table \ref{tab:Multiple Choice Datasets} shows the 
 availability of datasets in distractor generation tasks. 
 For example, CLOTH, DGen, SciQ, and MCQL are benchmark datasets in FITB and MC-QA tasks. Televic and EduQG are introduced specifically for distractor generation tasks. 
 RACE is a benchmark dataset in reading comprehension while two other datasets such as CosmosQA and DREAM are utilized in recent studies.  Visual7W  is the only multi-modal dataset used for textual distractor generation. Other datasets such as MedMCQA, MCTest, CBT, QuAIL and ReClor are utilized in the evaluation stage \cite{sharma2018using, wang2023weak, wang2023multi, wang2023efficient, ghanem2023disto, sileo-etal-2024-generating-multiple} for DG tasks.
 
 The majority of datasets are public except upon request datasets (e.g., SCDE, MovieQA) and upon payment of a license fee to access part of the dataset (e.g., WDW) or the whole dataset (e.g., Televic).

\section {Quantitative Results} \label{sec:quantitative_results}
\frenchspacing
The summary of quantitative results in DG tasks is detailed in this  section.

\subsection {Distractors in FITB and MC-QA} \label{sec:quantitative_results_FITB}
\frenchspacing
% The DG uses ranking-based metrics for FITB and MC-QA tasks, considering word-level distractors. 
Table \ref{tab:FITB-Results} summarizes the state-of-the-art (SOTA) results in DG for both FITB and MC-QA tasks, focusing on word-level distractors.
The most commonly used metric, precision P@1, yielded the following observations:
(i) retrieval-based methods utilizing feature-based learning outperformed neural networks based on adversarial training \cite{liang-etal-2018-distractor} in 
the 
SciQ\footnote{\citet{yu-etal-2024-enhancing} used ChatGPT to convert SciQ to FITB.}
and MCQL datasets;
(ii) context-aware neural networks fine-tuned with BERT \cite{bitew2022learning} achieved over 40\% relevant distractor retrieval in the Televic open-domain dataset;
(iii) SOTA results for the DGen and CLOTH datasets showed that fine-tuning Text2Text models with data augmentation strategies generated over 22\% relevant distractors.

% The most commonly used metric, precision P@1, was observed as follows:
% (i) retrieval based methods based on feature-based learning perform better than neural network based on adversarial training networks \cite{liang-etal-2018-distractor} in SciQ and MCQL datasets;
% (ii) retrieval methods based on context-aware neural network through BERT fine-tuning \cite{bitew2022learning} are able to retrieve relevant distractors more than 40\% in Televic dataset; 
% and (iii) the SOTA results for DGen and CLOTH datasets demonstrated successful fine-tuning of Text2Text models with data augmentation strategy to generate over 22\% relevant distractors.

\subsection {Distractors in MC-RC } \label{sec:quantitative_results-MC-RC}
\frenchspacing
% DG uses n-gram-based metrics for MC-RC tasks, considering word-level to sentence-level distractors. 
Table \ref{tab:MC-RC-Results} summarizes the SOTA results in MC-RC for DG using deep neural networks, focusing on word-level to sentence-level distractors. 
The collected studies used a RACE-modified dataset by \citet{gao2019generating}, excluding samples with distractors irrelevant to the passage and questions requiring option filling at the beginning or middle.
The most commonly used metric, BLUE, yielded the following observations: (i) The performance of the second and third distractors in beam search and multi-decoders showed a slight drop in BLEU-n scores due to lower likelihoods and a 0.5 Jaccard distance threshold, which enforced the use of different words. This drop was slightly less pronounced in MSG-Net due to its content selection approach. (ii) While the EDGE model achieved SOTA results in uni-gram matching for the three distractors, MSG-Net demonstrated the highest performance in bigram, trigram, and four-gram matching with the ground truth distractors.

In PLMs, \citet{chung-etal-2020-bert} fine-tuned the BERT model and achieved uni-gram, bigram, trigram, and four-gram matching scores of 39.81, 24.81, 17.66, and 13.56, respectively. The first distractors in fine-tuning T5 through two-step DG \cite{taslimipoor-etal-2024-distractor-generation} achieved uni-gram, bigram, trigram, and four-gram matching scores of 0.31,  0.20, 
0.15, and 0.12, respectively.

\begin{table*}[h]
\centering
% \footnotesize
\caption{Ranking-based metrics for DG in FITB and MC-QA tasks.}
\label{tab:FITB-Results}
\resizebox{0.77\textwidth}{!}{
\begin{tabular}{lccccc}
\hline\hline
Paper & Task & Dataset  & P@1 & NDCG@10 & MRR  \\ \hline
{LR+RF (\citeyear{liang-etal-2018-distractor})} & MC-QA  & SciQ   & \textbf{36.8} & 38.0 & 49.3  \\ 
{NN (\citeyear{liang-etal-2018-distractor})} & MC-QA  & SciQ   & 11.7 & 23.1 & 25.7  \\ 
%--- added
{RAP-T5 (\citeyear{yu-etal-2024-enhancing})} & FITB  & SciQ   & 24.30 & --- & 29.98  \\ \hline

{LR+RF (\citeyear{liang-etal-2018-distractor})} & MC-QA  & MCQL   &  \textbf{45.5} & 43.8  &  54.8  \\ 
{NN (\citeyear{liang-etal-2018-distractor})} & MC-QA  & MCQL   & 22.9 & 34.6 & 36.7  \\ \hline
{DQ-SIM (\citeyear{bitew2022learning})} & MC-QA  & Televic   &  \textbf{44.9} & --- &  62.8   \\ 
\hline

{EmbSim+CF  (\citeyear{jiang-lee-2017-distractor})} & FITB  & DGen  & 8.10 & 16.33  & 13.86  \\
{LR+RF (\citeyear{liang-etal-2018-distractor})} & FITB  & DGen  & 8.52 &  19.03& 15.87  \\ 
{BERT (\citeyear{devlin-etal-2019-bert})} & FITB  & DGen  & 7.72 & 16.21  &  13.60\\
{CSG-DS (\citeyear{ren2021knowledge})} & FITB  & DGen  & 10.85 & 19.70 & 17.51  \\ 
{CDGP (\citeyear{chiang-etal-2022-cdgp})} & FITB  & DGen  & 13.13 & 34.17  & 25.12 \\ 
{multi-task (\citeyear{wang-etal-2023-distractor})} & FITB  & DGen  & 22.00 & ---&  27.15  \\ 
{RAP-T5 (\citeyear{yu-etal-2024-enhancing})} & FITB  & DGen   & \textbf{22.39} & --- & 29.02  \\ \hline

\hline
{CDGP (\citeyear{chiang-etal-2022-cdgp})}   & FITB  & CLOTH & 18.50 & 37.82 & 29.96 \\ 
{multi-task (\citeyear{wang-etal-2023-distractor})} & FITB  & CLOTH   & \textbf{28.75} & --- & 34.46  \\ 
{two-step (\citeyear{taslimipoor-etal-2024-distractor-generation})}& FITB  & CLOTH & 26.57 & 47.29 & ---  \\ \hline
\end{tabular}
} 
\vspace{-2mm}
\end{table*}

\begin{table*}[h]
\centering
\footnotesize
\caption{N-gram metrics for DG using deep neural networks in MC-RC task within RACE dataset.}
\label{tab:MC-RC-Results}
\resizebox{0.8\textwidth}{!}{
\begin{tabular}{lcccccc}
\hline\hline
Paper   & Distractors  & BLEU-1 & BLEU-2 & BLEU-3 & BLEU-4  \\ \hline
HSA (\citeyear{gao2019generating})
 & \begin{tabular}[c]{@{}l@{}}1\textsuperscript{st}\\ 2\textsuperscript{nd}\\ 3\textsuperscript{rd}\end{tabular} & 
\begin{tabular}[c]{@{}l@{}}27.32 \\  26.56\\ 26.92\end{tabular}&
\begin{tabular}[c]{@{}l@{}}14.69 \\  13.14\\  12.88\end{tabular}& 
\begin{tabular}[c]{@{}l@{}}9.29\\  7.58\\  7.12\end{tabular}&
\begin{tabular}[c]{@{}l@{}}6.47\\  4.85\\  4.32\end{tabular}      \\ \hline

CHN (\citeyear{zhou2020co}) 
 & \begin{tabular}[c]{@{}l@{}}1\textsuperscript{st}\\ 2\textsuperscript{nd}\\ 3\textsuperscript{rd}\end{tabular} & 
\begin{tabular}[c]{@{}l@{}} 28.65\\  27.29\\  26.64\end{tabular}&
\begin{tabular}[c]{@{}l@{}} 15.15\\  13.57\\  12.67\end{tabular}&
\begin{tabular}[c]{@{}l@{}} 9.77\\  8.19\\  7.42\end{tabular}
& \begin{tabular}[c]{@{}l@{}} 7.01\\  5.51\\  4.88\end{tabular}              \\ \hline

EDGE (\citeyear{qiu-etal-2020-automatic})
 & \begin{tabular}[c]{@{}l@{}}1\textsuperscript{st}\\ 2\textsuperscript{nd}\\ 3\textsuperscript{rd}\end{tabular} & 
\begin{tabular}[c]{@{}l@{}} \textbf{33.03}\\ \textbf{32.07}\\ \textbf{31.29}\end{tabular}&
\begin{tabular}[c]{@{}l@{}}18.12\\ 16.75\\ 15.94\end{tabular}& 
\begin{tabular}[c]{@{}l@{}}11.35\\ 9.88\\ 9.24\end{tabular}&
\begin{tabular}[c]{@{}l@{}}7.57\\6.27\\ 5.70\end{tabular}     \\ \hline

HMD-Net (\citeyear{maurya2020learning})
 & \begin{tabular}[c]{@{}l@{}}1\textsuperscript{st}\\ 2\textsuperscript{nd}\\ 3\textsuperscript{rd}\end{tabular} & 
\begin{tabular}[c]{@{}l@{}}30.99 \\ 30.93\\ 29.70\end{tabular}&
\begin{tabular}[c]{@{}l@{}}17.30 \\ 16.89\\  15.95\end{tabular}&
\begin{tabular}[c]{@{}l@{}}11.09 \\ 10.64\\  9.74\end{tabular}& 
\begin{tabular}[c]{@{}l@{}}7.52 \\7.10\\  6.21\end{tabular}
\\ \hline

TMCA (\citeyear{shuai2021topic}) 
 & \begin{tabular}[c]{@{}l@{}}1\textsuperscript{st}\\ 2\textsuperscript{nd}\\ 3\textsuperscript{rd}\end{tabular} &  
\begin{tabular}[c]{@{}l@{}}29.01\\ 28.26\\ 27.18\end{tabular}&
\begin{tabular}[c]{@{}l@{}}14.84\\ 13.79\\ 12.55\end{tabular}&
\begin{tabular}[c]{@{}l@{}}9.61\\ 8.68\\ 7.64\end{tabular}&
\begin{tabular}[c]{@{}l@{}}6.87\\6.10\\ 5.04 \end{tabular} \\ \hline

MSG-Net (\citeyear{xie2021diverse}) 
 & \begin{tabular}[c]{@{}l@{}}1\textsuperscript{st}\\ 2\textsuperscript{nd}\\ 3\textsuperscript{rd}\end{tabular} & 
\begin{tabular}[c]{@{}l@{}}28.96\\ 27.91\\  27.84\end{tabular}&
\begin{tabular}[c]{@{}l@{}} \textbf{18.15}\\ \textbf{17.60}\\  \textbf{17.20}\end{tabular}& 
\begin{tabular}[c]{@{}l@{}}\textbf{12.31}\\ \textbf{12.26}\\  \textbf{11.81}\end{tabular}&
\begin{tabular}[c]{@{}l@{}}\textbf{8.87}\\ \textbf{8.86}\\  \textbf{8.53}\end{tabular}         \\ \hline
\end{tabular}
}
\end{table*}

\end{document}